\documentclass[10pt,twocolumn,letterpaper]{article}

\usepackage[pagenumbers]{cvpr} 

\definecolor{cvprblue}{rgb}{0.21,0.49,0.74}
\usepackage[pagebackref,breaklinks,hypertexnames=false,colorlinks,allcolors=cvprblue]{hyperref}
\usepackage{multirow}
\usepackage{pifont}

\def\confName{CVPR}
\def\confYear{2027}
\title{Exceeding Structural Pruning for Diffusion Transformers without Training}
\author{
    Peizhen Zhang$^{1}$\thanks{Equal contribution.} \quad
    Ziang Wu$^{2}$\footnotemark[1] \thanks{Work done during an internship at Tencent.} \quad
    Yang Li$^{1}$ \quad
    Shuai Shao$^{1}$ \quad
    Miles Yang$^{1}$ \quad
    Zhao Zhong$^{1}$\\[0.5em]
    {\normalsize $^{1}$Tencent Hunyuan} \quad
    {\normalsize $^{2}$Peking University}
}
\makeatletter
\g@addto@macro\@maketitle{
  \noindent\begin{minipage}{\textwidth}
    \centering
    \includegraphics[width=0.96\linewidth]{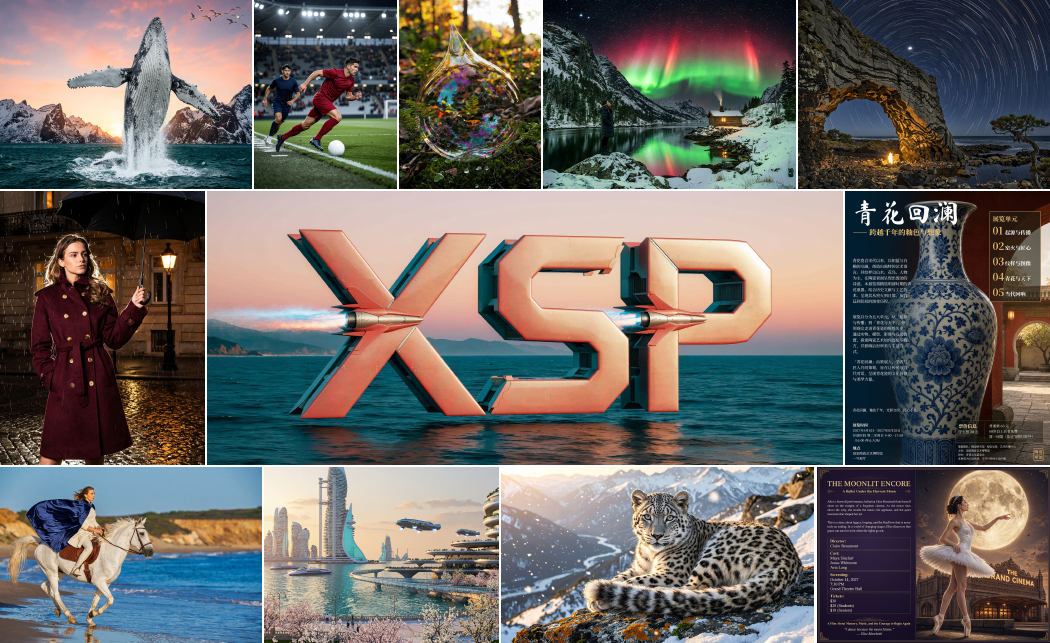}
    \captionof{figure}{Diverse 4K images generated by H3-XSP-16B, including the central XSP wordmark.}
    \label{fig:xsp-teaser}
  \end{minipage}\par
  \vspace{0.5em}
}
\makeatother
\begin{document}
\maketitle
\begin{abstract}
Modern generative models are getting better, which is normally accompanied by parameter up-scaling that gradually increases the deployment difficulty. Structural pruning is an effective solution to mitigate this issue but it often requires costly post-pruning recovery training (distillation or even further fine-tuning). To this end, we present XSP, a very efficient diffusion transformer-based structural compression framework. In this framework, we utilize a low-rank approximation to significantly reduce the AdaLN module parameters by 98.2\%. We propose a dynamic programming-based heterogeneous pruning architecture determination for FFNs. We devise a local least-squares reconstruction-based weight recovery for lightning-fast parameter adaptation for pruned attention heads and FFNs. Without regular gradient-descent-based recovery training which is necessary in existing methods, our method achieves the most competitive results under a pruning ratio of up to 50\%. On the Qwen-Image 1.0 pruning task, XSP ranks first on three out of five benchmarks. We also apply it to the recently open-sourced MiniMax-H3-Image (33B), where we obtain a 50\%-pruned 16B model with negligible quality drop and no performance degradation according to the benchmarks. We hope XSP establishes a strong baseline for the pruning research community.
\end{abstract}    
\section{Introduction}
\label{sec:intro}

In recent years, generative models of the diffusion paradigm dominate the image and video synthesis scenarios. Their backbones gradually evolve from convolutional neural network (CNN)-based to transformer-based. The prevalent architectures of diffusion transformers~\cite{peebles2023scalable} are friendly towards scaling and normally achieve good performance along with scaled-up parameters. This makes their deployment on cheap consumer-level GPUs much harder. Techniques such as structural pruning, sparse attention, and quantization are investigated to relieve this issue. In this paper, we focus on structural pruning and choose the popular image synthesis as the topic to analyze the potential redundancy for slimming. Without loss of generality, below, we start by analyzing Qwen-Image 1.0~\cite{qwenimage2025}, a classical text-to-image model, to gradually introduce our observations and inspirations step by step. We finally come up with a pruning framework that results in pruned models whose performance surpasses the previous SOTA without heavy recovery training surprisingly. Experimental results demonstrate that our method could be generalized to the recently released competitive image generation model MiniMax-H3-Image~\cite{minimax2026h3}.
\begin{figure}[t]
    \centering
    \includegraphics[width=\columnwidth]
        {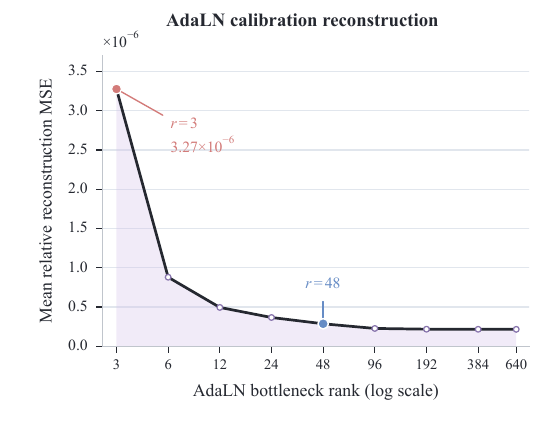}
    \caption{
        AdaLN reconstruction error versus bottleneck rank
        on Qwen-Image-1.0.
        Relative MSE is computed on the calibration timestep grid
        and averaged over 120 modulation projections.
        The horizontal axis uses a logarithmic scale.
        Reconstruction error increases as the rank decreases,
        particularly at the smallest ranks.
    }
    \label{fig:adaln_rank_error}
\end{figure}
The parameter distribution of the Qwen-Image 1.0 20B model consists of AdaLN (33\%), Attention (22\%) and FFN (45\%). In Qwen-Image 1.0, there are separate AdaLN projection modules across image/text throughout all blocks to modulate the timestep embeddings. Each module is mainly parameterized by a dense linear layer $\mathbb{R}^{d\times 6d}$ where $d$ reaches 3072 and 6 represents the shift, scale and gate outputs modulated for attention and FFN inputs. We hypothesize the redundant encoding and conduct exponential decay of low-rank approximation as shown in Fig.~\ref{fig:adaln_rank_error}. We see that the reconstruction MSE error is still below an extremely low level even when cutting 98\% of the AdaLN parameters. This naturally leads to the XSP-14B model specified in Table~\ref{tab:main_table}.
\begin{figure}[t]
    \centering
    \includegraphics[width=\columnwidth]{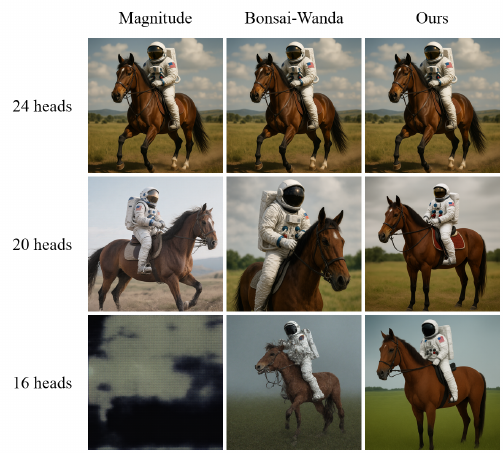}
    \caption{Attention head pruning showcases among strategies on Qwen-Image-1.0.
    Rows retain 24, 20, and 16 heads per block (the 24-head row serves as the non-pruned baseline and repeats); columns show head pruning choice by magnitude, attention head Wanda in Bonsai~\cite{dery2025bonsai} and our Wanda that generalizes GRAIL~\cite{tang2026grail} to dual-stream DiTs.
    }
    \label{fig:attention_head_ablation}
\end{figure}
Besides AdaLN, the attention mechanism determines the inter- and intra-token interaction and message exchange across modalities. We analyze its redundancy by dropping the attention heads which would simultaneously cut the output dimensions of the QKV matrices and the input dimension of the O matrix. As shown in Fig.~\ref{fig:attention_head_ablation}, the performance deterioration becomes evident as more heads are cut. To mitigate the degradation, we reconstruct the attention output projection via regularized least-squares fitting on calibration activations which will be specified in Sec.~\ref{mref:sec:attention}. Apparently, the phenomenon gets much relieved.
\begin{figure*}[t]
\centering
\includegraphics[width=\textwidth]{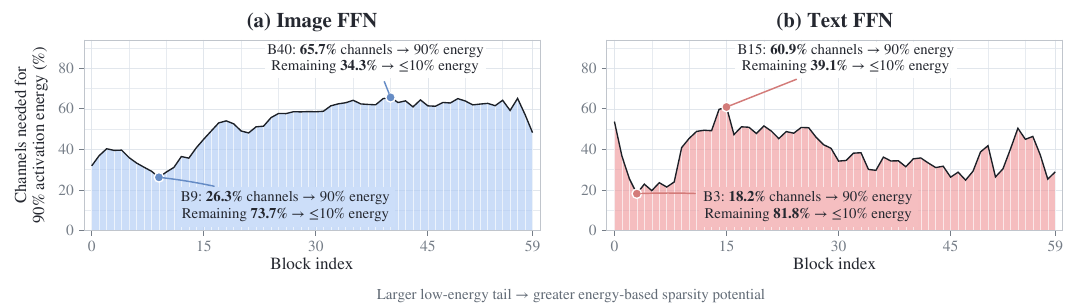}
\caption{\textbf{Uneven energy concentration motivates heterogeneous FFN widths.}
Each curve shows the minimum percentage of channels needed to capture 90\%
of activation energy in each FFN (image: blue; text: coral). Channels are
ranked by their mean squared activation before the output projection.
The callouts give the remaining channels, which contribute at most 10\%
of the energy. Statistics use 512 calibration examples at four noise
positions in the rank-48, 20-head Qwen-Image reference before FFN pruning
(12,288 channels per FFN).}
\label{mref:fig:ffn-energy}
\end{figure*}
As for FFN, as with attention pruning, we apply a similar strategy for uniform pruning of the linear layers throughout all transformer blocks. As in Table~\ref{tab:xsp_architectures}, it shows certain potential but it soon meets a bottleneck in the cutting ratio. However, the channel number in Qwen-1.0 FFN reaches 12,288, which may imply redundancy of different degrees. We conduct a heat map analysis as shown in Fig.~\ref{mref:fig:ffn-energy} to assess the energy during inference. By jointly considering the fact of non-uniform channel importance across layers and modalities and also the pruning budget constraint, we formulate the channel attainment problem as a multiple-choice knapsack problem. We use dynamic programming to resolve the channel pruning allocation which we specify in Sec.~\ref{mref:sec:ffn}. Fig.~\ref{fig:mckp_uniform_qualitative} demonstrates how such FFN heterogeneous width pruning pushes the limits of performance retention.
Our contributions are threefold:
\begin{itemize}
    \item We introduce XSP, an efficient framework for structured
    compression of diffusion transformers that \textbf{does not necessarily rely on
    fine-tuning or distillation}.
    \item We develop calibration-aware, layer-local AdaLN compression
    that preserves pretrained conditioning and modulation interfaces
    through low-rank reconstruction.
    \item We cast heterogeneous FFN capacity allocation as a
    reconstruction-aware constrained optimization problem, solved
    by dynamic programming to obtain globally optimal allocations
    for an additive sensitivity objective under an exact parameter budget.
\end{itemize}
\section{Related Work}
\label{sec:related_work}

\paragraph{Structured pruning and architecture selection.}
Importance-based pruning estimates which parameters, channels, or modules
can be removed with limited functional disruption. Structured importance
criteria and diffusion-specific sensitivity measurements have been studied
by Molchanov et al.~\citep{molchanov2019importance},
Diff-Pruning~\citep{fang2305structural},
LD-Pruner~\citep{castells2024ld}, and
OBS-DIFF~\citep{zhu2025obs}. Architectural compression of generative
transformers also includes approaches such as
SANA~\citep{xie2025sana} and PPCL~\citep{ma2025pluggable}.
XSP considers complementary structural dimensions within a fixed-depth
backbone and couples each selected structure to reconstructed weights.

\paragraph{Learned and timestep-dependent architectures.}
Differentiable masks, architecture search, and learned routing offer other
ways to allocate model capacity, as explored in
EcoDiff~\citep{zhang2024effortless},
TinyFusion~\citep{fang2025tinyfusion},
ALTER~\citep{yang2026alter}, and E-DiT~\citep{wang2026elastic}.
Related work also considers replacing operators, consolidating streams,
or varying capacity over the denoising trajectory, including
FastFLUX~\citep{cai2026fastflux},
Amber-Image~\citep{yang2026amber},
MosaicDiff~\citep{guo2025mosaicdiff}, and Diff-ES~\citep{liu2026diff}.
XSP does not train routing modules and does not switch architectures between
timesteps. Its head selections, ranks, and FFN widths are fixed before
inference.

\paragraph{Calibration and weight compensation.}
Training-free compression uses pretrained parameters and calibration
information to construct a compressed model without recovery training.
DiT-Pruning~\citep{hu2026posttrainingpruning} studies training-free
sparsification, while second-order methods such as
OBS-DIFF~\citep{zhu2025obs} motivate accounting for relationships between
feature dimensions. Other compression pipelines employ feature- or
prediction-level distillation, including TMP~\citep{zhang2026tmp},
HierarchicalPrune~\citep{kwon2026hierarchicalprune}, and
IGSM~\citep{zheng2025igsm}. XSP uses local reconstruction as the final
weight-compensation step. This reconstruction is distinct from fitting a
student through gradient-based feature or output distillation.

\paragraph{About AdaLN pruning.} Concurrently, there are two works applying AdaLN pruning to the video diffusion model MiniMax-H3~\cite{minimax2026h3}. One is FastH3 Trim~\cite{fasth3trim2026} by Hao AI Lab that leverages precomputed timestep modulation tables (or curve-fitted timestep representations). The other is HyperFlow H3~\cite{videorebirth2026hyperflowh3} by VideoRebirth. It utilizes a shared-conditioning dimensionality reduction but does not disclose enough detail. Our method independently factorizes each AdaLN projection using calibration-weighted SVD, ridge regularization, and bias correction as specified in Sec.~\ref{mref:sec:modulation}.
\begin{figure*}[t]
\centering
\includegraphics[width=0.97\textwidth]{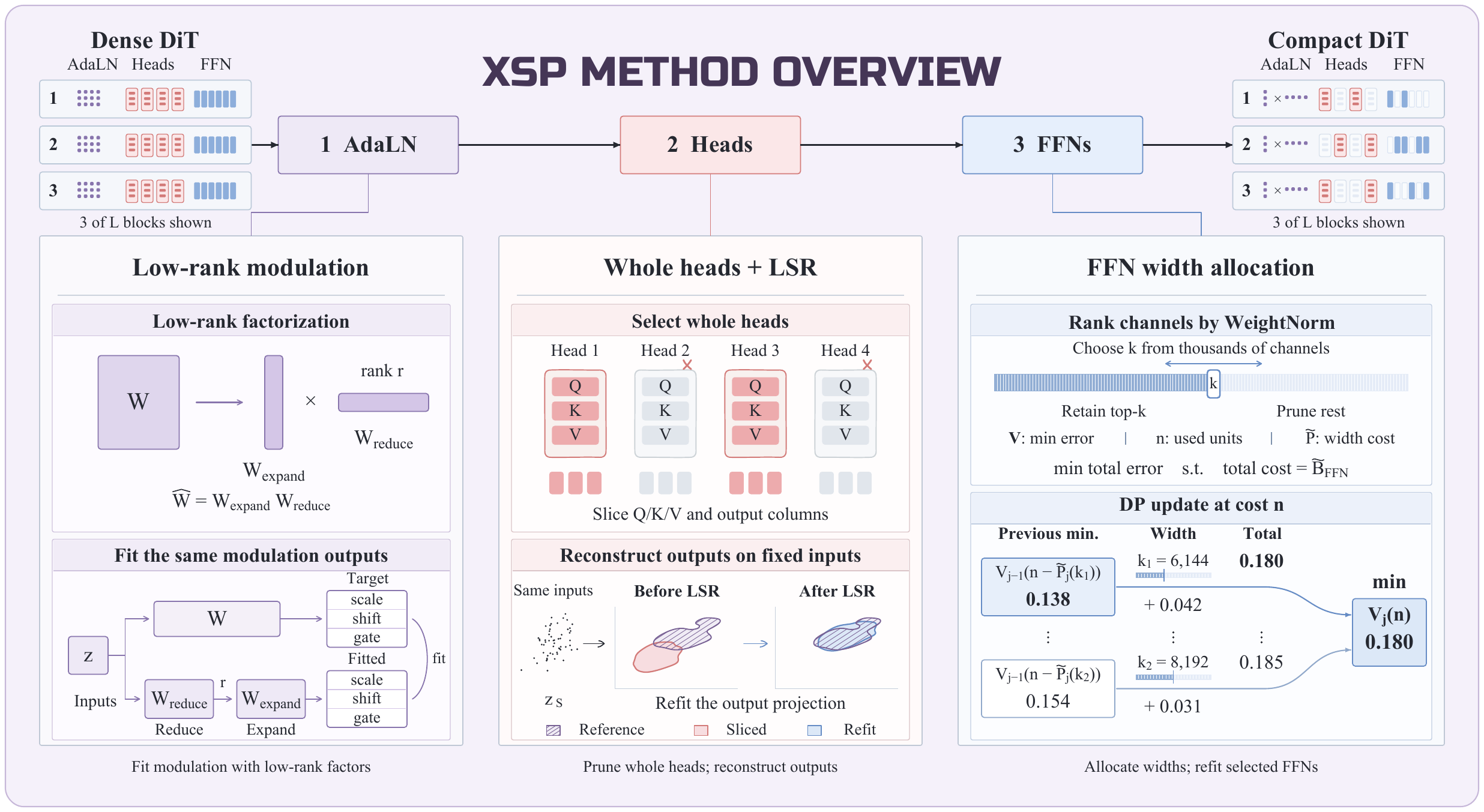}
\caption{\textbf{Overview of XSP.}
XSP compresses a DiT in three stages, preserving block count and residual dimensions.
\textbf{(1)~Low-rank modulation} fits low-rank factors to each AdaLN projection on calibration inputs.
\textbf{(2)~Whole-head pruning + LSR} removes full attention heads and refits the output projection on retained features via least-squares reconstruction (LSR).
\textbf{(3)~FFN width allocation} ranks channels by WeightNorm to form LSR-refitted candidates and allocates widths by dynamic programming under the shared remaining FFN budget.}
\label{mref:fig:overview}
\end{figure*}
\paragraph{About Attention Head pruning.} Bonsai~\cite{dery2025bonsai} is an LLM-oriented compression solution. It involves attention head pruning starting from structured Wanda~\cite{sun2024wanda} priors, sampling masked submodels, and fitting a mask-to-performance regression to refine head importance. It iteratively prunes low-scoring structures and updates the priors. The regression improves selection, without reconstructing the surviving weights. Our strategy directly ranks heads by aggregating output-weight column L2 norms multiplied by activation RMS, replacing structured Wanda’s absolute-weight averaging. After selecting heads, full-Gram least-squares reconstruction adjusts the retained output projection, using channel correlations to compensate for removed heads. Bonsai emphasizes search-based selection; ours combines direct selection with explicit weight reconstruction. Wanda-SP in GRAIL~\cite{tang2026grail} proposes a Wanda-like + LSR paradigm. Our method differs from theirs in the weighting manner between activations and weights, regularization choice and generalization towards the DiT scenario: GRAIL uses L1-based scoring and zero-centered ridge regularization. We score heads using activation-weighted output-column L2 norms and regularize reconstructed weights towards the retained original weights. Ours generalizes to dual-stream DiTs where we additionally combine stream-normalized scores to select shared heads, then reconstruct each stream separately, preserving native head dimensions and residual interfaces.
\section{Methodology}
\label{mref:sec:method}

\subsection{Problem Formulation and Overview}
\label{mref:sec:overview}

We formulate structural compression of a diffusion transformer as finding a budget-constrained subnetwork whose predictions match the pretrained model across denoising trajectories. Let $M_0$ be the
pretrained model, $M$ a compressed model, and $P(M)$ its transformer parameter
count. The prediction $f_M(x_t,t,c)$ takes a noisy latent $x_t$, timestep $t$,
and conditioning $c$. Given an input distribution $\mathcal D$ and budget
$B$, the compression objective is
\begin{equation}
\begin{aligned}
\min_{M\in\mathcal S(M_0)}\quad &
\mathbb E_{\mathcal D}
\|f_M(x_t,t,c)-f_{M_0}(x_t,t,c)\|_2^2\\
\text{subject to}\quad &P(M)\leq B.
\end{aligned}
\label{mref:eq:objective}
\end{equation}
Here $\mathcal S(M_0)$ contains models with compressed AdaLN projections,
attention heads, and FFN channels, preserving the block sequence and residual
dimensions. Their structural choices are the AdaLN ranks $\mathbf r=(r_m)$,
retained head counts $\mathbf h=(h_l)$, and FFN widths $\mathbf w=(w_j)$;
$m$, $l$, and $j$ index the respective modules. Separate modality branches
are separate FFN modules. Throughout, $B$ denotes a parameter budget,
$P$ denotes parameter counts, and $E$ denotes candidate reconstruction
errors. For local linear mappings, we use $z$ for input features, $W$ for
weights, $\beta$ for bias, and $y=Wz+\beta$ for the reference output;
hats indicate reconstructed quantities. Calibration examples
$\mathcal D_{\mathrm{cal}}$ supply fitting statistics, and a disjoint probe
set $\mathcal D_{\mathrm{probe}}$ measures candidate errors.

\paragraph{Decoupled Surrogate and Staged Pipeline.}
Directly optimizing Eq.~\eqref{mref:eq:objective} is intractable, as $\mathcal{S}(M_0)$ spans an exponentially large combinatorial space and evaluating each candidate requires full-model denoising forwards.
To make this tractable, we decouple module parameterization from cross-layer budget allocation: candidate modules are first parameterized by fitting their reference outputs on $\mathcal{D}_{\mathrm{cal}}$, and FFN widths are then allocated according to block-output errors on $\mathcal{D}_{\mathrm{probe}}$.
Specifically, output fitting adjusts output projections in attention and FFNs so retained features compensate for pruned ones, while approximating AdaLN projections via low-rank mappings over conditioning activations.

As shown in Figure~\ref{mref:fig:overview}, XSP executes this construction sequentially across three checkpoints: AdaLN factorization yields $M_{\mathrm{mod}}$, whole-head pruning with reconstruction yields $M_{\mathrm{attn}}$, and budgeted FFN allocation produces the final model $M_\star$.
Before each stage, calibration statistics and reconstruction targets are refreshed from the preceding checkpoint and held fixed within the stage.
Given a configuration $(\mathbf r, \mathbf h)$, all width-independent parameters $P_{\mathrm{fixed}}(\mathbf r, \mathbf h)$ are determined, leaving an exact budget $B_{\mathrm{FFN}} = B - P_{\mathrm{fixed}}(\mathbf r, \mathbf h)$ that directly couples attention head retention with available FFN capacity.
Crucially, we allocate $B_{\mathrm{FFN}}$ using the block errors remaining \emph{after} reconstruction, coupling all three stages through sequential output preservation under a shared parameter budget.
\subsection{Activation-Aware Low-Rank Modulation Compression}
\label{mref:sec:modulation}

AdaLN maps conditioning embeddings to the scales, shifts, and gates that
modulate transformer features. Although these projections have large output
dimensions, their inputs can occupy a much smaller effective subspace.
We therefore preserve the directions used by conditioning activations when
fitting a smaller mapping.

For a projection $y=Wz+\beta$ of size $d_{\mathrm{out}}\times d_{\mathrm{in}}$
and target rank $r$, we fit
\begin{equation}
\min_{\operatorname{rank}(\widehat W)\leq r,\,\widehat\beta}
\mathbb E_{\mathrm{cal}}\|y-\widehat Wz-\widehat\beta\|_2^2
+\lambda_{\mathrm{mod}}\|W-\widehat W\|_F^2.
\label{mref:eq:mod-objective}
\end{equation}
The output error weights directions by how strongly they are activated;
the regularizer stabilizes fitting in weakly observed directions.
This objective is solved by a covariance-weighted truncated SVD.
The resulting factors $W_{\mathrm{reduce}}\in\mathbb R^{r\times d_{\mathrm{in}}}$
and $W_{\mathrm{expand}}\in\mathbb R^{d_{\mathrm{out}}\times r}$ implement
$\widehat W=W_{\mathrm{expand}}W_{\mathrm{reduce}}$ as two consecutive linear
maps. Their weight count is $r(d_{\mathrm{in}}+d_{\mathrm{out}})$ instead of
$d_{\mathrm{in}}d_{\mathrm{out}}$, while the modulation outputs retain their
original dimensions and roles. Appendix~A.1 gives
the factorization and bias correction.

This stage compresses the conditioning pathway without changing the
attention or FFN interfaces. The resulting checkpoint supplies the
activations used to select and reconstruct attention heads.

\subsection{Whole-Head Attention Pruning}
\label{mref:sec:attention}

Attention compression reduces the features used for token interaction.
We remove complete heads, together with their query, key, value, and output
projection groups. Keeping each retained head's dimension preserves its
positional encoding geometry and produces smaller dense attention operations.

\paragraph{Select heads by their output contribution.}
A head matters when its features are active and the output projection gives
those features substantial weight. Let $z^{(s)}$ be the concatenated head
outputs in modality stream $s$, $W^{(s)}$ their output projection, and
$\mathcal I_a$ the feature indices of head $a$. We score the head as
\begin{equation}
\rho_{a}^{\mathrm{head},(s)}=
\sum_{i\in\mathcal I_a}\|W_{:,i}^{(s)}\|_2
\sqrt{\mathbb E_{\mathrm{cal}}[(z_i^{(s)})^2]}.
\label{mref:eq:head-score}
\end{equation}
For coupled streams, we normalize scores by each stream's mean and sum
them before retaining the top $h_l$ heads in attention module $l$.
The selected head indices determine all corresponding projection slices;
joint attention shares this selection across its modality branches.

\paragraph{Recover the output from retained features.}
Simply slicing the projections loses the contribution of removed heads.
We compensate by refitting the output projection to predict the reference
output from the retained features. For retained feature indices $S$, this
least-squares reconstruction (LSR) solves
\begin{equation}
\min_{\widehat W,\widehat\beta}\quad
\mathbb E_{\mathrm{cal}}\|\widehat Wz_S+\widehat\beta-y\|_2^2
+\lambda_{\mathrm{rec}}\|\widehat W-W_{:,S}\|_F^2.
\label{mref:eq:lsr}
\end{equation}
The first term restores the output using the surviving features, while the
second anchors the fit to the corresponding pretrained weights.
The same operation will reconstruct FFN candidates in the next stage:
only the retained feature set changes. The closed-form solution and
numerical details appear in Appendix~A.2.

After reconstructing attention, $M_{\mathrm{attn}}$ defines the reference
for FFN compression. Its head configuration also determines the remaining
FFN budget, so fewer heads permit wider FFNs at the same total model size.

\subsection{Exact-Budget Heterogeneous Width Allocation}
\label{mref:sec:ffn}

The remaining budget need not be distributed uniformly. As
Figure~\ref{mref:fig:ffn-energy} shows, capturing 90\% of activation energy
requires 26.3--65.7\% of channels across image FFNs and 18.2--60.9\% across
text FFNs. The same retained width therefore covers different amounts of
energy in different modules. To turn this heterogeneity into a width
allocation, we compare how well each FFN can recover its output at several
widths. Capacity is then assigned according to the error left after recovery.

\paragraph{Build and reconstruct candidate widths.}
For FFN $j$, let $\mathcal K_j$ be its candidate widths. At each width
$k\in\mathcal K_j$, we retain the top-$k$ intermediate channels ranked by
the product of their associated input- and output-weight norms.
This ranks the complete path through a channel, including both gate and up
projections in a gated FFN. We slice the coupled projection rows and columns,
then refit the output projection using the same LSR objective as attention
(Eq.~\eqref{mref:eq:lsr}). Each candidate thus has a concrete parameter count
$P_j(k)$ and its own reconstructed weights. The ranking rule is specified
in Appendix~A.3.

\paragraph{Measure the error that remains.}
We insert each reconstructed candidate into its reference block, changing
only FFN $j$, and compare block outputs on identical probe inputs.
Measuring at the block output includes the effect of gating and residual
connections on the candidate's contribution. We use the squared output
difference relative to the reference output energy, averaged over probe
examples and output streams. For separate modality branches, we additionally
normalize by each branch's mean error at its smallest candidate widths and
apply the chosen branch weights. Denote this allocation error by $E_j(k)$;
Appendix~A.4 gives its precise definition.
The resulting table pairs parameter count $P_j(k)$ with error $E_j(k)$
for every candidate. It separates channel selection within a width from
the decision of how much width each FFN should receive.

\paragraph{Choose widths within the shared budget.}
We select one candidate per FFN to minimize the sum of these measured errors:
\begin{equation}
\begin{aligned}
\min_{\mathbf w}\quad &\sum_j E_j(w_j)\\
\text{subject to}\quad &w_j\in\mathcal K_j,\qquad
\sum_j P_j(w_j)=B_{\mathrm{FFN}}.
\end{aligned}
\label{mref:eq:mckp}
\end{equation}
This multiple-choice knapsack problem is solved by dynamic programming across all $J$ FFN modules. For exact-budget comparisons, the candidate sets include a feasible uniform allocation at the same parameter count. Let $\Delta$ be a common integer parameter unit dividing all candidate costs and the remaining budget without rounding, giving scaled integer costs $\widetilde P_j(k)=P_j(k)/\Delta$ and budget $\widetilde B_{\mathrm{FFN}}=B_{\mathrm{FFN}}/\Delta$. Let $V_j(n)$ denote the minimum summed candidate error over the first $j$ FFNs at exact cost $n$, satisfying the recurrence
\begin{equation}
V_j(n)=\min_{\substack{k\in\mathcal K_j\\\widetilde P_j(k)\leq n}}
\left[V_{j-1}(n-\widetilde P_j(k))+E_j(k)\right].
\label{mref:eq:dp-main}
\end{equation}
We initialize $V_0(0)=0$ and $V_0(n)=+\infty$ for $n>0$; states with no feasible transition also take infinite error. Storing each state's minimizing width and backtracking from $V_J(\widetilde B_{\mathrm{FFN}})$ recovers the optimal allocation for this additive objective. We then assemble the chosen channel slices and reconstructed projections into $M_\star$. Because allocation uses residual errors measured after LSR, an FFN can receive fewer channels when its lost contribution is more readily recovered, redistributing capacity across modules.

\subsection{Adapting the Framework to Various Backbones}
\label{mref:sec:backbones}

Applying XSP to a backbone requires identifying its modulation projections,
coupled attention heads, and FFN channel groups. These determine which
parameters must be retained together and the cost of each candidate.
Joint attention shares head selections across coupled streams; standard
FFNs couple input rows with output columns, while gated FFNs additionally
couple the gate rows. Separate image and text FFNs can receive different
widths within the same allocation problem.

For example, Qwen-Image provides separate image and text FFN allocation
units in each block, with shared head selections for joint attention.
Appendix~A.6 gives its dimensions and channel costs.
Across these layouts, the output-preserving fits and budgeted allocation
produce physically smaller dense modules with the original residual
interfaces. All parameters are initialized from pretrained weights and
calibration statistics, without gradient-based fine-tuning or distillation.

\section{Experiments}
\label{sec:xsp_experiments}

\begin{table*}[t]
\caption{Comparison to existing pruning methods across commonly used benchmarks on Qwen-Image 1.0. R.(\%) represents the average performance drop rate, a legacy metric from PPCL.
Overall, XSP-pruned models achieve most of the best performances on the Qwen-Image 1.0 pruning task. We also report results on the recent state-of-the-art open-source model MiniMax-H3-Image to show the generalization ability of XSP. Notably, our pruned 16B model still behaves on par with the original 33B model.}
\label{tab:main_table}
\centering
\renewcommand{\arraystretch}{1.12}
\setlength{\tabcolsep}{4pt}
\resizebox{\textwidth}{!}{
\begin{tabular}{c|c|c|c|c|ccccccccc|c}
\hline
\multirow{2}{*}{Models}
& \multirow{2}{*}{Methods}
& \multirow{2}{*}{Params. (B)}
& \multirow{2}{*}{KD Recovery}
& \multirow{2}{*}{Post-KD FT}
& \multirow{2}{*}{DPG$\uparrow$}
& \multirow{2}{*}{GenEval$\uparrow$}
& \multicolumn{2}{c}{LongText$\uparrow$}
& \multicolumn{2}{c}{OneIG$\uparrow$}
& \multicolumn{3}{c|}{T2I-CompBench$\uparrow$}
& \multirow{2}{*}{R.(\%)$\downarrow$} \\
\cline{8-14}
& & & & & & & EN & ZH & EN & ZH
& B-VQA & UniDet & S-CoT & \\
\hline
\multirow{9}{*}{Qwen-Image 1.0}
& Base model
& 20 & - & - & 88.9 & 0.870
& 0.943 & 0.946 & 0.539 & 0.548
& 0.709 & 0.532 & 82.47 & 0 \\
\cline{2-15}
& TinyFusion
& 14 & \ding{52} & \ding{56} & 80.7 & 0.739
& 0.859 & 0.857 & 0.500 & 0.498
& 0.689 & 0.464 & 78.99 & 8.75 \\
& HierarchicalPrune
& 14 & \ding{52} & \ding{56} &  83.3 & 0.766
& 0.884 & 0.881 & 0.509 & 0.496
& 0.706 & 0.487 & \underline{79.94} & 6.49 \\
& PPCL
& 14 & \ding{52} & \ding{56} & \underline{87.9} & \underline{0.847}
& \underline{0.929} & \textbf{0.946} & \textbf{0.536} & \textbf{0.538}
& \underline{0.750} & \underline{0.524} & \textbf{82.15} & 0.42 \\
& \textbf{XSP} (Ours)
& 14 & \ding{56} & \ding{56} & \textbf{89.4}  & \textbf{0.866}
& \textbf{0.940} & \underline{0.944} & \underline{0.531} & \underline{0.521}
& \textbf{0.751} & \textbf{0.543} & 79.8 & \textbf{0.23} \\
\cline{2-15}
& PPCL
& 10 & \ding{52} & \ding{56} &  81.7 & 0.784
& 0.871 & 0.885 & 0.485 & 0.501
& 0.701 & 0.499 & 79.24 & 6.88 \\
& PPCL-FT
& 10 & \ding{52} & \ding{52} & 86.7 & 0.828
& 0.902 & \underline{0.931} & 0.499 & \textbf{0.519}
& 0.715 & 0.515 & \textbf{81.33} & 3.29 \\
& Amber-Image
& 10 & \ding{52} & \ding{52} & \textbf{89.6}  & \textbf{0.881}
& \underline{0.911} & 0.915 & \underline{0.504} & 0.502
& - & - & - & -  \\
& \textbf{XSP} (Ours)
& 10 & \ding{56} & \ding{56} & \underline{88.4} & \underline{0.847}
& \textbf{0.938} & \textbf{0.948} & \textbf{0.535} & \textbf{0.519}
& \textbf{0.741} & \textbf{0.551} & \underline{79.77} & \textbf{0.52} \\
\hline
\multirow{3}{*}{MiniMax-H3-Image}
& Base model & 33 & - & - & 85.5 & 0.827
& 0.979 & 0.963 & 0.541  & 0.503
& 0.724 & 0.496 & 79.5 & 0 \\
\cline{2-15} 
& \textbf{XSP} (Ours)& 20 & \ding{56} & \ding{56} & 85.5 & 0.827
& 0.979 & 0.962 & 0.541 & 
 0.503 & 0.725 & 0.495 & 79.5 & 0.02 \\
& \textbf{XSP} (Ours)& 16 & \ding{56} & \ding{56} & 85.6 & 0.836
& 0.976 & 0.963 & 0.538 & 0.500
& 0.723 & 0.496 & 79.5 & 0.03 \\
\hline
\end{tabular}
}

\end{table*}

\subsection{Setup}

\paragraph{Models and benchmarks.} We experiment on Qwen-Image-1.0 (50 steps) and MiniMax-H3-Image (6 steps) to perform adequate verification of our pruning framework. The former is one of the most widely adopted models for pruning studies. The latter is the recently open-sourced SOTA model. We primarily conduct ablation studies on Qwen-Image-1.0 without loss of generality. In our experiments, every XSP-pruned model is evaluated directly after compression, without undergoing regular recovery training such as distillation (feature or output) and fine-tuning. During inference, we retain the denoising hyper-parameters (sigmas, resolution and classifier-free guidance scale, etc.) of the original models for their pruned variants. We use five common benchmarks to sufficiently validate the effectiveness of our method and conduct comparisons to existing methods: DPG~\cite{hu2024ella}, GenEval~\cite{ghosh2023geneval}, LongText~\cite{xomni2025}, OneIG~\cite{oneig2025bench}, and T2I-CompBench~\cite{huang2023t2icompbench}.
 
\paragraph{Inference and scoring.} For XSP Qwen-Image-1.0 image pruning experiments, images are all generated at $1328\times1328$ in BF16 using 50 denoising steps and batch size 1. CFG is 4 for LongText and 8 for the remaining benchmarks. Benchmark prompts, sample IDs, and seed assignments are frozen within each evaluation protocol. Model-specific prompt suffixes are identical across variants, the negative prompt is a single blank, and
the maximum text sequence length is 512. LongText uses 160 prompts per language with seeds $\{1111,1234,4321,8888\}$, yielding 640 images per language. We follow the established benchmark-specific scoring procedures. In our summary, B-VQA averages color, shape, and texture; UniDet averages 2D spatial, 3D spatial, and numeracy; and OneIG averages its five component scores. Non-spatial CLIP scoring is reported separately from UniDet.

\paragraph{Calibration and probing in XSP.}
Unless otherwise specified, this setup applies to Qwen-Image-1.0 pruning. We use 512 calibration and 128 disjoint probe examples from a fixed internal image--caption pool. Under fixed noisy-latent replay, we form noisy inputs for each example by interpolating with Gaussian noise at four fixed noise positions, using cached unconditional conditioning on every tenth example--timestep pair. We fit rank-48 AdaLN projections on a 4K uniformly spaced timestep grid, and collect attention and FFN moments from evenly sampled valid tokens to reconstruct projections in FP64. Finally, we probe reconstructed FFN candidates for budget allocation.

\subsection{Comparison to Existing Methods}

As shown in Table~\ref{tab:main_table}, for Qwen-Image 1.0, we provide 14B and 10B pruned variants of XSP. XSP-14B represents the AdaLN-pruned model and XSP-10B further considers joint attention and FFN pruning (20-head and mean-width-8192 MCKP). Notably, compared to previous methods, our model achieves first place on three to four out of five benchmarks, using neither knowledge-distillation-based recovery training nor fine-tuning. Both 14B and 10B variants reach the lowest average performance drop relative to the original model, compared to their competing counterparts, respectively. This undoubtedly establishes a strong baseline for future research. We also provide results on MiniMax-H3-Image. The 20B version stands for the AdaLN-pruned version while the 16B variant further considers joint attention and FFN pruning (48-head and mean-width-11136 MCKP). Notably, there is no performance degradation of the 50\%-pruned 16B model against the original model. On GenEval, it even improves over the original model (+0.9\%) perhaps due to less over-fitting.

\begin{table}[t]
\caption{GenEval scores ($\uparrow$) for attention-head-only pruning
on Qwen-Image-1.0. AdaLN and FFNs remain unchanged.
}
\label{tab:attention_head_ablation}
\centering
\renewcommand{\arraystretch}{1.15}
\setlength{\tabcolsep}{6pt}
\begin{tabular}{c|ccc}
\hline
\multicolumn{4}{c}{Original model (24 heads): 0.870} \\
\hline
Heads & Magnitude & Bonsai-Wanda & Ours \\
\hline
20 & 0.812 & 0.848 & \textbf{0.859} \\
16 & 0.023 & 0.375 & \textbf{0.797} \\
\hline
\end{tabular}

\end{table}

\subsection{Ablation Studies}

\paragraph{AdaLN pruning study.}
As shown in Figure~\ref{fig:adaln_rank_error}, evidently, once the AdaLN bottleneck rank exceeds 48, the reconstruction error flattens out, reaching a consistently low plateau. The rank-48 model corresponds to the XSP-14B model for Qwen-Image 1.0. As shown in Table~\ref{tab:main_table}, it performs comparably to the original 20B model. Likewise, for the MiniMax-H3-Image model, we use rank 16 for AdaLN pruning by a similar analysis, which leads to the XSP-20B model in the same table. Unless otherwise specified, the pruned variants in the ablation studies below all go through AdaLN pruning.

\paragraph{Attention head pruning study.}

Beforehand, we show visualization evidence of our Wanda advantage in Figure~\ref{fig:attention_head_ablation} for attention head pruning. We show results on GenEval for these architectures in Table~\ref{tab:attention_head_ablation}.

\paragraph{FFN pruning study.}
We experiment on an FFN-pruned-only (uniformly) 10B model derived from Qwen-Image-1.0 without loss of generality.
Table~\ref{tab:geneval_channel_lsr} shows that ranking channels by magnitude and then applying LSR is the best practice.

\begin{figure}[!t]
  \centering
  \includegraphics[width=\linewidth]{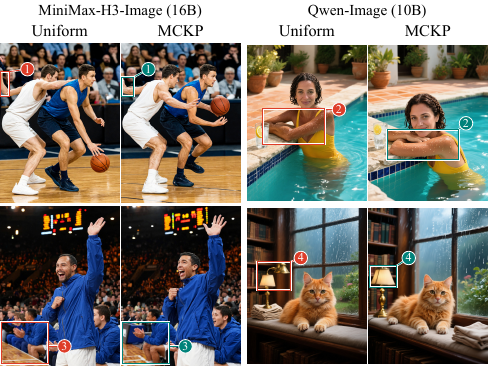}
  \caption{Uniform versus MCKP pruning on MiniMax-H3-Image (left,
  16B) and Qwen-Image (right, 10B), with matched prompts and seeds
  within each pair. Boxes mark artifacts in uniform outputs absent from
  their MCKP counterparts: a malformed hand (1), an unnatural arm
  connection (2), a raised floor/bench boundary (3), and a duplicated
  lampshade (4).}
  \label{fig:mckp_uniform_qualitative}
\end{figure}

\paragraph{Heterogeneous FFN pruning by MCKP.}
Quantitatively, Table~\ref{tab:xsp_architectures} shows steady improvement of MCKP (+0.04\%, +2.9\% and +4.9\%) against uniform FFN channel pruning across different settings of 10B pruned variants for Qwen-Image 1.0. Qualitatively, Figure~\ref{fig:mckp_uniform_qualitative} shows the mitigation of structural artifacts and enhancement of visual quality and fine-grained details of MCKP against uniform pruning.

\paragraph{Joint attention-FFN capacity trade-offs.}
As aforementioned, Table~\ref{tab:xsp_architectures} gives three 10B joint pruning
configurations (rank-48 AdaLN and either uniform or MCKP-allocated
FFN widths). Uniform and heterogeneous models within the same row have exactly the
same parameter count. The 20-head and 8192-Avg-FFN-Width configuration attains the highest observed score. It refers to the 10B pruned model in Table~\ref{tab:main_table}. 

\begin{table}[t]
\caption{Ablations for FFN pruning hyper-parameters (Magnitude \textit{vs.} Wanda, w/ and w/o LSR). We conduct the study on an FFN-pruned-only architecture where all 24 attention heads are kept and FFN width gets uniformly pruned to reach 10B parameters.}
\label{tab:geneval_channel_lsr}
\centering
\small
\setlength{\tabcolsep}{3pt}
\begin{tabular}{c|c|cc|cc|c}
\hline
\multirow{2}{*}{Heads}
& \multirow{2}{*}{FFN width}
& \multicolumn{2}{c|}{Channel choice}
& \multicolumn{2}{c|}{LSR}
& \multirow{2}{*}{GenEval$\uparrow$} \\
\cline{3-6}
& & Magnitude & Wanda & w/o & w/ & \\
\hline
\multirow{4}{*}{24}
& \multirow{4}{*}{7168}
& \ding{51} &           & \ding{51} &           & 0.0000 \\
& & \ding{51} &           &           & \ding{51} & \textbf{0.8227} \\
& &           & \ding{51} & \ding{51} &           & 0.2342 \\
& &           & \ding{51} &           & \ding{51} & 0.8081 \\
\hline
\end{tabular}

\end{table}

\begin{table}[t]
\caption{Ablations (GenEval$\uparrow$) for trading off joint pruning of attention heads and average FFN width for achieving 10B pruning. The Avg. FFN width implies that the FFNs across layers and branches could either be uniformly cut or heterogeneously cut by our proposed MCKP method. These variants all apply the best practice for attention head and channel pruning mentioned above.}
\label{tab:xsp_architectures}
\centering
\small
\setlength{\tabcolsep}{5pt}
\begin{tabular}{cccc}
\toprule
\multirow{2}{*}{Heads}
& \multirow{2}{*}{Avg. FFN width}
& \multicolumn{2}{c}{Pruned FFN Architecture} \\
\cmidrule(lr){3-4}
& & Uniform & MCKP \\
\midrule
24 & 7168 & 0.8227 & 0.8231 \\
20 & 8192 & 0.8186 & \textbf{0.8471} \\
16 & 9216 & 0.7286 & 0.7779 \\
\bottomrule
\end{tabular}

\end{table}

\subsection{Scope and Limitations}

XSP focuses on training-free structural compression of pretrained diffusion transformers. The applicable compression operators and structural constraints depend on the target backbone, and extending the framework to additional architectures may require corresponding adaptations. As a calibration-based approach, its reconstruction and capacity allocation are influenced by the calibration distribution, so different application domains may benefit from different calibration mixtures. Our findings therefore apply to the evaluated backbones and generation settings, rather than establishing a universally optimal compression configuration across all diffusion models and applications.

\section{Conclusion}
We presented XSP, a training-free framework for structural compression of diffusion transformers. XSP combines architecture-aware compression with calibration-based reconstruction, supporting low-rank modulation projections, whole-head attention pruning, and budget-constrained heterogeneous FFN width allocation. By preserving residual interfaces, it produces compact, static architectures without recovery training or distillation. Experiments across multiple backbones demonstrate favorable compression–quality trade-offs, while ablations highlight the importance of weight reconstruction and capacity allocation. These findings establish XSP as a practical approach to reducing the model footprint of large diffusion transformers and motivate further exploration of calibration strategies and architecture-specific compression choices. We hope that XSP can become one of the new paradigms for DiT pruning within the community.
\pagebreak
\clearpage
\section*{LLM Usage Statement}
ChatGPT and Gemini were used to refine the wording and improve the clarity of the manuscript.
\par

{
    \small
    \bibliographystyle{ieeenat_fullname}
    \bibliography{main}
}
\clearpage
\begingroup
\raggedbottom
\setcounter{page}{1}
\setcounter{figure}{0}
\setcounter{table}{0}
\numberwithin{equation}{section}
\renewcommand{\thefigure}{S\arabic{figure}}
\renewcommand{\thetable}{S\arabic{table}}
\renewcommand{\eqref}[1]{\textup{(\ref*{#1})}}
\hypersetup{hidelinks}
\setcounter{dbltopnumber}{2}
\renewcommand{\dbltopfraction}{0.9}
\renewcommand{\textfraction}{0.08}
\renewcommand{\dblfloatpagefraction}{0.75}
\setlength{\dbltextfloatsep}{10pt}
\makeatletter
\def\@arrayrule{\@addtopreamble{\vline}}
\makeatother

\maketitlesupplementary
\appendix
\noindent This supplement explains how XSP is constructed and evaluated, what structure it retains, and how it pairs with recovery training and quantization. Section~\ref{mref:app:details} provides derivations and implementation details for modulation factorization, least-squares reconstruction, and dynamic programming allocation. Sections~\ref{supp:sec:t2i-details} and~\ref{supp:sec:editing} expand the T2I benchmark breakdowns and test direct transfer to instruction-guided image editing. Section~\ref{supp:sec:structural-analysis} examines FFN allocation, channel selection, and attention head reconstruction, then connects this preserved structure to end-to-end generation quality before student recovery training. Section~\ref{supp:extreme_prune_ratio_tiny_prune} evaluates pruning and lightweight distillation recovery on the compact Z-Image Turbo model, and Section~\ref{supp:sec:quantization} tests compatibility with post-training quantization. Section~\ref{supp:sec:qualitative} collects visual comparisons in the same order as the experiments: T2I, editing, then quantization.
\begin{center}
\begin{minipage}{\columnwidth}
\captionof{table}{Key XSP notation.}
\label{supp:tab:notation}
\small
\renewcommand{\arraystretch}{1.05}
\begin{tabular}{@{}p{0.35\linewidth}p{0.59\linewidth}@{}}
\toprule
\textbf{Symbol} & \textbf{Meaning} \\
\midrule
$r_m,h_l,w_j$ & Rank, head, and width choices \\
$B, B_{\mathrm{FFN}}$ & Total and FFN parameter budgets \\
$\mathcal K_j$ & Candidate FFN widths \\
$P_j(k)$ & Candidate parameter count \\
$E_j(k)$ & Candidate allocation error \\
$\Delta$ & Integer parameter unit \\
$n$ & Parameter units used \\
$V_j(n)$ & Minimum summed candidate error \\
\bottomrule
\end{tabular}
\par\smallskip{\footnotesize\emph{Equation references.} Supplementary equations carry section-letter prefixes; unprefixed equation numbers refer to the main paper.}
\end{minipage}
\end{center}

\begingroup
\linespread{0.94}\selectfont
\AtBeginEnvironment{equation}{
\setlength{\abovedisplayskip}{4pt}
\setlength{\belowdisplayskip}{4pt}
\setlength{\abovedisplayshortskip}{2pt}
\setlength{\belowdisplayshortskip}{2pt}
}
\setlength{\abovedisplayskip}{5pt}
\setlength{\belowdisplayskip}{5pt}
\setlength{\abovedisplayshortskip}{3pt}
\setlength{\belowdisplayshortskip}{3pt}
\section{Derivations and Implementation Details for XSP}
\label{mref:app:details}

\subsection{Activation-Aware Modulation Factorization}
\label{mref:app:modulation}

For the local mapping $y=Wz+\beta$, let $\mu=\mathbb E_{\mathrm{cal}}[z]$
and $C=\mathbb E_{\mathrm{cal}}[(z-\mu)(z-\mu)^\top]$.
Optimizing the bias in Eq.~\eqref{mref:eq:mod-objective} gives
$\widehat\beta=\beta+(W-\widehat W)\mu$.
Let $R$ be a covariance factor satisfying
$RR^\top=C+\lambda_{\mathrm{mod}}I$. The remaining objective is
$\|(W-\widehat W)R\|_F^2$.
With the leading rank-$r$ singular components of $WR$ written as
$U_r\Sigma_rV_r^\top$, the solution is
\begin{equation}
\begin{aligned}
W_{\mathrm{expand}}&=U_r\Sigma_r^{1/2},\\
W_{\mathrm{reduce}}&=\Sigma_r^{1/2}V_r^\top R^{-1},\\
\widehat W&=W_{\mathrm{expand}}W_{\mathrm{reduce}},\\
\widehat\beta&=\beta+(W-\widehat W)\mu.
\end{aligned}
\label{mref:eq:mod-fit}
\end{equation}
We approximate the truncated decomposition with randomized SVD and compute
the bias correction from the factors after casting them to the export
precision. The two factors are consecutive linear maps with no intervening
activation. For a bias-free projection, we use the uncentered second moment
$\mathbb E_{\mathrm{cal}}[zz^\top]$ in place of $C$ and keep the bias zero.

\subsection{Head Aggregation and Shared Output Reconstruction}
\label{mref:app:reconstruction}

For coupled attention streams, let $\overline\rho^{\mathrm{head},(s)}$
be the mean of the head scores in stream $s$. The aggregate score is
\begin{equation}
\rho_a^{\mathrm{head}}=
\sum_s\frac{\rho_a^{\mathrm{head},(s)}}
{\max(\overline\rho^{\mathrm{head},(s)},\epsilon_h)}.
\end{equation}
We retain the top $h_l$ heads in module $l$ and share their indices across
the coupled projection groups.

For reconstruction, $S$ denotes the retained feature indices and
$\overline S$ their complement. Using the input mean $\mu$ and covariance
$C$ defined above for the current local mapping, the solution to
Eq.~\eqref{mref:eq:lsr} is
\begin{equation}
\begin{aligned}
\widehat W&=W_{:,S}+W_{:,\overline S}
C_{\overline S S}(C_{SS}+\lambda_{\mathrm{rec}}I)^{-1},\\
\widehat\beta&=\beta+W\mu-\widehat W\mu_S.
\end{aligned}
\label{mref:eq:lsr-solution}
\end{equation}
The correction uses correlations between removed and retained features to
recover their contribution through the retained projection. For a bias-free
mapping, we use uncentered second moments and keep the intercept zero.
The regularization strength scales with the mean diagonal energy of the
retained features, with a positive floor. We solve the linear systems in
FP64. After casting to the export precision, we accept the fitted projection
when its calibration reconstruction MSE is no larger than that of direct
slicing and its norm inflation is bounded; otherwise we retain the sliced
projection. Attention and FFN reconstruction use this same procedure.

\subsection{FFN Channel Ranking and Candidate Costs}
\label{mref:app:candidates}

Let $\mathcal W_j^{\mathrm{in}}$ be the input-side weight matrices of FFN
$j$, and $W_j^{\mathrm{out}}$ its output projection. The channel score is
\begin{equation}
\rho_{j,i}^{\mathrm{ffn}}=
\|W_{j,:,i}^{\mathrm{out}}\|_2
\prod_{W\in\mathcal W_j^{\mathrm{in}}}\|W_{i,:}\|_2.
\label{mref:eq:channel-score}
\end{equation}
A standard FFN has one input-side matrix; a gated FFN has both up and gate
matrices. For width $k$, the top-$k$ channel indices specify the input rows
and output columns retained together. These channel sets are nested as
$k$ increases, but each candidate has its own fitted output projection.

$P_j(k)$ counts the width-dependent parameters of candidate $k$, including
any biases that change with the retained width. Width-independent biases
belong to $P_{\mathrm{fixed}}(\mathbf r,\mathbf h)$, so
$P_{\mathrm{fixed}}+\sum_jP_j(w_j)$ is the total transformer parameter count.

\subsection{Block-Output Errors and Branch Normalization}
\label{mref:app:probe}

To score a candidate, we replace only FFN $j$ in its enclosing block of
$M_{\mathrm{attn}}$. Let $y_j^{(s)}$ and $\widehat y_{j,k}^{(s)}$ be
reference and candidate block outputs on identical inputs in stream $s$.
Let $\mathcal T_j$ contain the block's output streams. Define
\begin{equation}
\overline E_j(k)=\mathbb E_{\mathrm{probe}}\left[
\frac{1}{|\mathcal T_j|}\sum_{s\in\mathcal T_j}
E_{\mathrm{rel}}(\widehat y_{j,k}^{(s)},y_j^{(s)})\right],
\label{mref:eq:probe}
\end{equation}
where
\begin{equation}
E_{\mathrm{rel}}(\widehat y,y)=
\begin{cases}
\|\widehat y-y\|_F^2/\|y\|_F^2,&\|y\|_F^2\geq\epsilon_E,\\
0,&\text{otherwise}.
\end{cases}
\label{mref:eq:relative-error}
\end{equation}
Both energies use valid tokens only. The block boundary includes gating
and residual operations, so the error measures their effect on the
reconstructed FFN contribution.

For modality-specific FFNs, let $\mathcal J_s$ be the modules in branch
$s$ and $s(j)$ the branch of module $j$. We normalize by the branch's mean
error at the smallest candidate widths:
\begin{equation}
\begin{aligned}
\alpha_s&=\max\left\{
\frac{1}{|\mathcal J_s|}\sum_{j\in\mathcal J_s}
\overline E_j(\min\mathcal K_j),\epsilon_\alpha\right\},\\
E_j(k)&=\omega_{s(j)}\overline E_j(k)/\alpha_{s(j)}.
\end{aligned}
\label{mref:eq:normalization}
\end{equation}
Here $\omega_s$ is the branch weight. This produces the allocation errors
used in Eq.~\eqref{mref:eq:mckp}; without branch normalization, the raw
candidate error $\overline E_j(k)$ can be used directly.

\subsection{Exact-Budget Dynamic Programming}
\label{mref:app:allocation}

Candidate widths follow the chosen channel granularity. For the matched
uniform and heterogeneous comparisons, every candidate set includes an
admissible uniform width whose total cost defines $B_{\mathrm{FFN}}$.
This ensures that the equality constraint in Eq.~\eqref{mref:eq:mckp}
has a feasible solution.

Let $\Delta$ be an integer parameter unit dividing all candidate costs and
the remaining budget, and define
$\widetilde P_j(k)=P_j(k)/\Delta$ and
$\widetilde B_{\mathrm{FFN}}=B_{\mathrm{FFN}}/\Delta$.
For $J$ FFNs, let $V_j(n)$ be the minimum summed candidate error for the
first $j$ modules using exactly $n$ parameter units. Its recurrence is
\begin{equation}
V_j(n)=\min_{\substack{k\in\mathcal K_j\\\widetilde P_j(k)\leq n}}
\left[V_{j-1}(n-\widetilde P_j(k))+E_j(k)\right].
\label{mref:eq:dp}
\end{equation}
Initialize $V_0(0)=0$ and all other states to infinity. Backtracking from
$V_J(\widetilde B_{\mathrm{FFN}})$ recovers the selected widths.
The optimized quantity is the sum of single-module probe errors;
the assembled model is evaluated in the experiments.

\subsection{Backbone Dimensions and Channel Costs}
\label{mref:app:backbones}

Qwen-Image 1.0 has 60 blocks with separate image and text FFNs, giving
120 allocation units. Its modulation projections are fitted individually
using their shared timestep-input moments. At rank 48, each projection
becomes $3072\rightarrow48\rightarrow18432$, preserving all six native
modulation outputs. Joint attention retains common head indices across
branches, with head dimension 128 and residual width 3072.

For a standard GELU FFN with residual width $d_{\mathrm{res}}$ and an
input bias, a channel couples one input-projection row, one bias entry,
and one output-projection column, costing $2d_{\mathrm{res}}+1$ parameters.
A bias-free gated FFN couples a gate row, an up row, and a down column,
costing $3d_{\mathrm{res}}$ parameters per channel. Shared query/key/value
groups are treated as coupled pruning units, and modulation statistics
include the conditioning variables used by the native mapping.

\endgroup

\makeatletter
\renewcommand\paragraph{\@startsection{paragraph}{4}{\z@}{0.8ex plus .2ex}{-1em}{\normalfont\normalsize\bfseries}}
\makeatother
\section{Detailed Text-to-Image Evaluation}

\label{supp:sec:t2i-details}

We expand the original-model and XSP rows of main-paper Table~\ref*{tab:main_table} into benchmark components. These are the same evaluations used in the main table; the original Qwen-Image-1.0 baseline follows the published PPCL results (Ma et al., 2025).

\paragraph{Generation and sample coverage.}

Qwen evaluations use $1328\times1328$ images and 50 denoising steps. MiniMax-H3-Image uses official prompt expansion (PE), with one cached rewrite shared across variants, and six PDD steps at approximately 16 million pixels. PE selects the aspect ratio; height and width are multiples of 32. Scoring uses the original benchmark prompts and annotations. Prompt IDs and seeds are fixed within each protocol; LongText uses seeds 1111, 1234, 4321, and 8888 for each language.

\begin{table}[!t]
\centering
\caption{Evaluation coverage for XSP variants and the original H3 model. Repeats are images per prompt.}
\label{supp:tab:t2i-protocol}
\small
\setlength{\tabcolsep}{4pt}
\renewcommand{\arraystretch}{1.08}
\begin{tabular}{@{}lrrr@{}}
\toprule
Benchmark & Prompts & Repeats & Images \\
\midrule
DPG & 1,065 & 4 & 4,260 \\
GenEval & 553 & 4 & 2,212 \\
OneIG EN & 1,120 & 4 & 4,480 \\
OneIG ZH & 1,320 & 4 & 5,280 \\
CompBench (8 tasks) & $8\times300$ & 10 & 24,000 \\
LongText (each language) & 160 & 4 & 640 \\
\bottomrule
\end{tabular}
\end{table}

\begin{table*}[!t]

\centering

\captionsetup{skip=6pt}

\caption{DPG component scores (0--100). Categories use dependency-filtered question outcomes; Overall averages per-image means. Published Qwen categories use a different aggregation and are left blank.}
\label{supp:tab:t2i-dpg}
\begingroup
\small
\setlength{\tabcolsep}{3pt}
\renewcommand{\arraystretch}{1.10}
\begin{tabular*}{\linewidth}{@{}l@{\extracolsep{\fill}}rrrrrr@{}}
\toprule
Variant & Global & Entity & Attribute & Relation & Other & Overall \\
\midrule
\multicolumn{7}{@{}l}{\emph{Qwen-Image 1.0}} \\
Original (20B)$^\dagger$ & --- & --- & --- & --- & --- & 88.9 \\
XSP-14B & 83.7 & 91.9 & 87.6 & 89.2 & 87.0 & 89.4 \\
XSP-10B & 81.1 & 91.2 & 86.6 & 87.5 & 85.2 & 88.4 \\
\midrule
\multicolumn{7}{@{}l}{\emph{MiniMax-H3-Image}} \\
Original (33B) & 84.5 & 88.8 & 82.7 & 84.7 & 81.0 & 85.5 \\
XSP-20B & 83.7 & 88.7 & 82.7 & 84.7 & 81.7 & 85.5 \\
XSP-16B & 82.4 & 88.6 & 82.8 & 84.5 & 81.3 & 85.6 \\
\bottomrule
\end{tabular*}
\endgroup

\medskip

\caption{GenEval category accuracies. Overall is the mean of the six categories.}
\label{supp:tab:t2i-geneval}
\begingroup
\small
\setlength{\tabcolsep}{3pt}
\renewcommand{\arraystretch}{1.10}
\begin{tabular*}{\linewidth}{@{}l@{\extracolsep{\fill}}rrrrrrr@{}}
\toprule
Variant & \shortstack{Single\\object} & \shortstack{Two\\objects} & Counting & Colors & Position & \shortstack{Attribute\\binding} & Overall \\
\midrule
\multicolumn{8}{@{}l}{\emph{Qwen-Image 1.0}} \\
Original (20B)$^\dagger$ & 0.99 & 0.92 & 0.89 & 0.88 & 0.76 & 0.77 & 0.870 \\
XSP-14B & 1.000 & 0.972 & 0.884 & 0.867 & 0.713 & 0.758 & 0.866 \\
XSP-10B & 0.991 & 0.980 & 0.831 & 0.880 & 0.630 & 0.770 & 0.847 \\
\midrule
\multicolumn{8}{@{}l}{\emph{MiniMax-H3-Image}} \\
Original (33B) & 0.988 & 0.917 & 0.803 & 0.835 & 0.798 & 0.620 & 0.827 \\
XSP-20B & 0.988 & 0.927 & 0.806 & 0.822 & 0.798 & 0.620 & 0.827 \\
XSP-16B & 0.988 & 0.932 & 0.822 & 0.838 & 0.798 & 0.640 & 0.836 \\
\bottomrule
\end{tabular*}
\endgroup

\end{table*}

\begin{table*}[!t]

\centering

\captionsetup{skip=6pt}

\caption{OneIG English: the five component scores and their equally weighted Overall.}
\label{supp:tab:t2i-oneig-en}
\begingroup
\small
\setlength{\tabcolsep}{3pt}
\renewcommand{\arraystretch}{1.10}
\begin{tabular*}{\linewidth}{@{}l@{\extracolsep{\fill}}rrrrrr@{}}
\toprule
Variant & Alignment & Text & Reasoning & Style & Diversity & Overall \\
\midrule
\multicolumn{7}{@{}l}{\emph{Qwen-Image 1.0}} \\
Original (20B)$^\dagger$ & 0.882 & 0.891 & 0.306 & 0.418 & 0.197 & 0.539 \\
XSP-14B & 0.889 & 0.947 & 0.291 & 0.393 & 0.136 & 0.531 \\
XSP-10B & 0.870 & 0.967 & 0.293 & 0.391 & 0.151 & 0.535 \\
\midrule
\multicolumn{7}{@{}l}{\emph{MiniMax-H3-Image}} \\
Original (33B) & 0.881 & 0.951 & 0.348 & 0.402 & 0.121 & 0.541 \\
XSP-20B & 0.880 & 0.953 & 0.348 & 0.401 & 0.121 & 0.541 \\
XSP-16B & 0.881 & 0.951 & 0.348 & 0.390 & 0.121 & 0.538 \\
\bottomrule
\end{tabular*}
\endgroup

\medskip

\caption{OneIG Chinese, using the same five components and aggregation as English.}
\label{supp:tab:t2i-oneig-zh}
\begingroup
\small
\setlength{\tabcolsep}{3pt}
\renewcommand{\arraystretch}{1.10}
\begin{tabular*}{\linewidth}{@{}l@{\extracolsep{\fill}}rrrrrr@{}}
\toprule
Variant & Alignment & Text & Reasoning & Style & Diversity & Overall \\
\midrule
\multicolumn{7}{@{}l}{\emph{Qwen-Image 1.0}} \\
Original (20B)$^\dagger$ & 0.825 & 0.963 & 0.267 & 0.405 & 0.279 & 0.548 \\
XSP-14B & 0.836 & 0.982 & 0.275 & 0.378 & 0.132 & 0.521 \\
XSP-10B & 0.815 & 0.989 & 0.272 & 0.372 & 0.149 & 0.519 \\
\midrule
\multicolumn{7}{@{}l}{\emph{MiniMax-H3-Image}} \\
Original (33B) & 0.841 & 0.843 & 0.315 & 0.389 & 0.125 & 0.503 \\
XSP-20B & 0.841 & 0.845 & 0.315 & 0.389 & 0.125 & 0.503 \\
XSP-16B & 0.842 & 0.847 & 0.316 & 0.370 & 0.125 & 0.500 \\
\bottomrule
\end{tabular*}
\endgroup

\end{table*}

\begin{table*}[!t]

\centering

\captionsetup{skip=6pt}

\caption{CompBench components. B-VQA averages color, shape, and texture; UniDet averages spatial, 3D spatial, and numeracy. Non-spatial CLIP is separate. S-CoT is on a 0--100 scale.}
\label{supp:tab:t2i-compbench}
\begingroup
\footnotesize
\setlength{\tabcolsep}{3pt}
\renewcommand{\arraystretch}{1.10}
\begin{tabular*}{\linewidth}{@{}l@{\extracolsep{\fill}}rrrrrrrrrr@{}}
\toprule
& \multicolumn{4}{c}{B-VQA} & \multicolumn{4}{c}{UniDet} & & \\
\cmidrule(lr){2-5}\cmidrule(lr){6-9}
Variant & Color & Shape & Texture & Avg. & Spatial & 3D & Numeracy & Avg. & \shortstack{Non-\\spatial} & S-CoT \\
\midrule
\multicolumn{11}{@{}l}{\emph{Qwen-Image 1.0}} \\
Original (20B)$^\dagger$ & --- & --- & --- & 0.709 & 0.428 & 0.453 & 0.716 & 0.532 & --- & 82.47 \\
XSP-14B & 0.853 & 0.624 & 0.775 & 0.751 & 0.421 & 0.461 & 0.746 & 0.543 & 0.313 & 79.8 \\
XSP-10B & 0.838 & 0.617 & 0.767 & 0.741 & 0.429 & 0.470 & 0.754 & 0.551 & 0.316 & 79.77 \\
\midrule
\multicolumn{11}{@{}l}{\emph{MiniMax-H3-Image}} \\
Original (33B) & 0.826 & 0.634 & 0.713 & 0.724 & 0.442 & 0.400 & 0.646 & 0.496 & 0.306 & 79.5 \\
XSP-20B & 0.825 & 0.634 & 0.714 & 0.725 & 0.439 & 0.401 & 0.647 & 0.495 & 0.306 & 79.5 \\
XSP-16B & 0.823 & 0.636 & 0.710 & 0.723 & 0.438 & 0.398 & 0.652 & 0.496 & 0.305 & 79.5 \\
\bottomrule
\end{tabular*}
\endgroup

\medskip

\caption{LongText OCR matching rates. Overall pools matched and ground-truth counts across short and long prompts.}
\label{supp:tab:t2i-longtext}
\begingroup
\small
\setlength{\tabcolsep}{3pt}
\renewcommand{\arraystretch}{1.10}
\begin{tabular*}{\linewidth}{@{}l@{\extracolsep{\fill}}rrrrrr@{}}
\toprule
& \multicolumn{3}{c}{English} & \multicolumn{3}{c}{Chinese} \\
\cmidrule(lr){2-4}\cmidrule(lr){5-7}
Variant & Short & Long & Overall & Short & Long & Overall \\
\midrule
\multicolumn{7}{@{}l}{\emph{Qwen-Image 1.0}} \\
Original (20B)$^\dagger$ & --- & --- & 0.943 & --- & --- & 0.946 \\
XSP-14B & 0.947 & 0.937 & 0.940 & 0.938 & 0.946 & 0.944 \\
XSP-10B & 0.938 & 0.938 & 0.938 & 0.935 & 0.954 & 0.948 \\
\midrule
\multicolumn{7}{@{}l}{\emph{MiniMax-H3-Image}} \\
Original (33B) & 0.958 & 0.988 & 0.979 & 0.976 & 0.958 & 0.963 \\
XSP-20B & 0.960 & 0.987 & 0.979 & 0.979 & 0.955 & 0.962 \\
XSP-16B & 0.953 & 0.986 & 0.976 & 0.972 & 0.959 & 0.963 \\
\bottomrule
\end{tabular*}
\endgroup

\end{table*}

\paragraph{Scoring and aggregation.}

Tables~\ref{supp:tab:t2i-dpg}--\ref{supp:tab:t2i-longtext} report the component scores. DPG uses mPLUG with dependency filtering: categories average question outcomes, whereas Overall averages per-image question means. GenEval averages six category accuracies.

OneIG averages five dimensions: alignment and text use Qwen2.5-VL, reasoning uses LLM2CLIP, style uses CSD and StyleEncoder, and diversity averages DreamSim over the six pairs formed by four images.

For CompBench, B-VQA averages color, shape, and texture; UniDet averages spatial, 3D spatial, and numeracy. Non-spatial CLIP is separate, and ShareGPT4V supplies S-CoT on a 0--100 scale. LongText pools matched and ground-truth counts across short and long prompts before dividing, rather than averaging subset rates.

\paragraph{Reading the component results.}
The breakdown distinguishes capabilities that an overall score can obscure. Qwen XSP-10B retains high single- and two-object accuracy ($0.991/0.980$), while counting and position fall to $0.831/0.630$ from the published $0.89/0.76$ baseline. Its OneIG text scores increase while diversity decreases in both languages. H3-16B stays close across GenEval categories ($0.836$ overall versus $0.827$ for the original model).

\noindent $^\dagger$ denotes published Qwen original-model scores at their reported precision. DPG category scores use a different published aggregation and are left blank; other dashes indicate unavailable components. Overall scores match the main table.

\section{Cross-Task Transfer to Instruction-Guided Image Editing}
\label{supp:sec:editing}
We test whether a structure chosen using only T2I calibration also supports instruction-guided image editing. The same XSP-16B checkpoint is applied directly, without editing-specific calibration, fine-tuning, or recovery training, and compared with the original H3-33B model.

\begin{table*}[!t]
\centering
\caption{Instruction-guided image editing comparison on MiniMax-H3-Image. For GEdit, $G_{\mathrm{SC}}$, $G_{\mathrm{PQ}}$, and $G_{\mathrm{O}}$ denote semantic consistency, perceptual quality, and overall scores from the GPT-4.1 judge ($G$); higher is better for all metrics.}
\label{supp:tab:editing}
\fontsize{9}{10.5}\selectfont
\setlength{\tabcolsep}{3.5pt}
\renewcommand{\arraystretch}{1.08}
\begin{tabular*}{\textwidth}{@{\extracolsep{\fill}}lrrrrrrrr@{}}
\toprule
 & \multicolumn{3}{c}{\shortstack{GEdit-CN (0--10)$^{\dagger}$\\$N=605$}}
 & \multicolumn{3}{c}{\shortstack{GEdit-EN (0--10)\\$N=606$}}
 & \shortstack{ImgEdit (1--5)\\$N=737$}
 & \shortstack{KRIS (0--100)\\$N=1{,}267$} \\
\cmidrule(lr){2-4}\cmidrule(lr){5-7}\cmidrule(lr){8-8}\cmidrule(lr){9-9}
\textbf{Model} & $G_{\mathrm{SC}}$ & $G_{\mathrm{PQ}}$ & $G_{\mathrm{O}}$
 & $G_{\mathrm{SC}}$ & $G_{\mathrm{PQ}}$ & $G_{\mathrm{O}}$ & Overall & Overall \\
\midrule
H3-33B (original) & 9.14 & 8.24 & 8.52 & 9.11 & 8.14 & 8.46 & 4.59 & 87.36 \\
XSP-16B (pruned) & 8.99 & 8.20 & 8.38 & 9.04 & 8.14 & 8.42 & 4.60 & 87.99 \\
\bottomrule
\end{tabular*}
\par\smallskip
\begin{minipage}{\textwidth}
\footnotesize
$^{\dagger}$GEdit overall (O) aggregates per-image $\sqrt{\mathrm{SC}\cdot\mathrm{PQ}}$ with all 11 categories equally weighted. On GEdit-CN, GPT-4.1 refused perceptual-quality scoring for one original H3-33B output; this same example is omitted for both models ($605/606$ evaluated, without score imputation).
\end{minipage}
\end{table*}

\paragraph{Generation Setup}
For each example, both models receive the same reference images, expanded instruction, and seed (42). We use the released MiniMax-H3-Image-PE weights with the official editing task prompt and message assembly. Both models sample in six PDD steps at roughly 16 million pixels, using the PE-selected aspect ratio and dimensions aligned to 32, with CFG 1 and no negative prompt.

\paragraph{Evaluation Protocol}
We follow the BAGEL evaluation protocols for GEdit (CN and EN), ImgEdit, and KRIS. GEdit uses GPT-4.1 (2025-04-14); ImgEdit and KRIS use GPT-4o (2024-11-20). Instruction-conditioned judges receive the original benchmark instruction, while quality-only subjudges receive no text instruction. Image preprocessing and score aggregation follow each benchmark's protocol. The shared GEdit-CN exclusion is explained below Table~\ref{supp:tab:editing}.

\paragraph{Quantitative Results}
At roughly half the transformer parameter count, XSP-16B remains close to H3-33B on all three benchmarks (Table~\ref{supp:tab:editing}). Both semantic consistency and perceptual quality remain high in the two GEdit languages. This comparison supports transfer from T2I calibration to editing without an editing-specific adaptation stage.

\section{Structural Pruning Analysis}

\label{supp:sec:structural-analysis}

\subsection{FFN Allocation and Channel Selection}

\begin{figure*}[!t]

\centering
\includegraphics[width=0.85\linewidth]{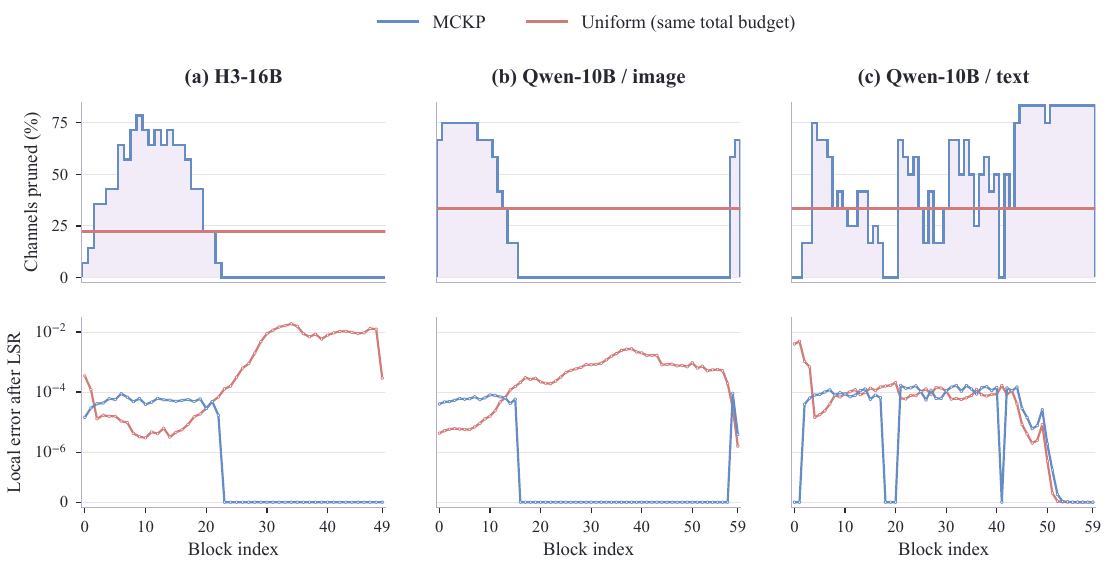}
\caption{FFN allocation at matched total budget $B_{\mathrm{FFN}}$: H3-16B (left, 50 FFNs; $2{,}048$ fit/$512$ probe examples) and Qwen-10B (middle/right, 60 image/60 text FFNs; $512$ fit/$128$ probe). Top: fraction of channels pruned by MCKP (blue, lavender fill) and uniform pruning (coral). Bottom: raw local block-output errors $\overline E_j(w_j)$ after LSR from the same probe table, including full-width modules. The shared log scale is linear below $10^{-7}$ to display zeros.}
\label{supp:fig:ffn-allocation-analysis}

\end{figure*}

Here, we clarify the origin of MCKP (short for the multiple-choice knapsack problem) referenced in the main text: we formulate heterogeneous channel pruning as a multiple-choice knapsack problem and solve it via dynamic programming, abbreviating this solver as MCKP. Figure~\ref{supp:fig:ffn-allocation-analysis} compares MCKP and uniform allocation at identical transformer sizes: $16.486$B for H3-16B and $9.984$B for Qwen-10B. With AdaLN ranks and head counts fixed, both use the same candidate probe table and FFN budget $\sum_jP_j(w_j)=B_{\mathrm{FFN}}$, with $w_j\in\mathcal K_j$ (Section~\ref{mref:app:allocation}).

H3 starts from width $14{,}336$: uniform pruning uses $11{,}136$ everywhere, while MCKP spans $3{,}072$--$14{,}336$ and leaves blocks 23--49 full-width. Qwen starts from $12{,}288$, with uniform width $8{,}192$. MCKP averages $10{,}069.3$ image and $6{,}314.7$ text channels; image blocks 16--57 remain full-width, while late text blocks mostly use $2{,}048$.

The lower panels show raw local errors $\overline E_j(w_j)$. Using the allocation errors $E_j$ from Section~\ref{mref:app:probe}, including Qwen's branch normalization, MCKP reduces $\sum_jE_j(w_j)$ by $99.48\%$ on H3 and $90.69\%$ on Qwen. Uneven widths therefore lower the additive probe objective at the same budget; these reductions measure local reconstruction, not benchmark gains.

\begin{figure*}[!t]

\captionsetup{skip=4pt}

\centering
\includegraphics[width=0.80\linewidth]{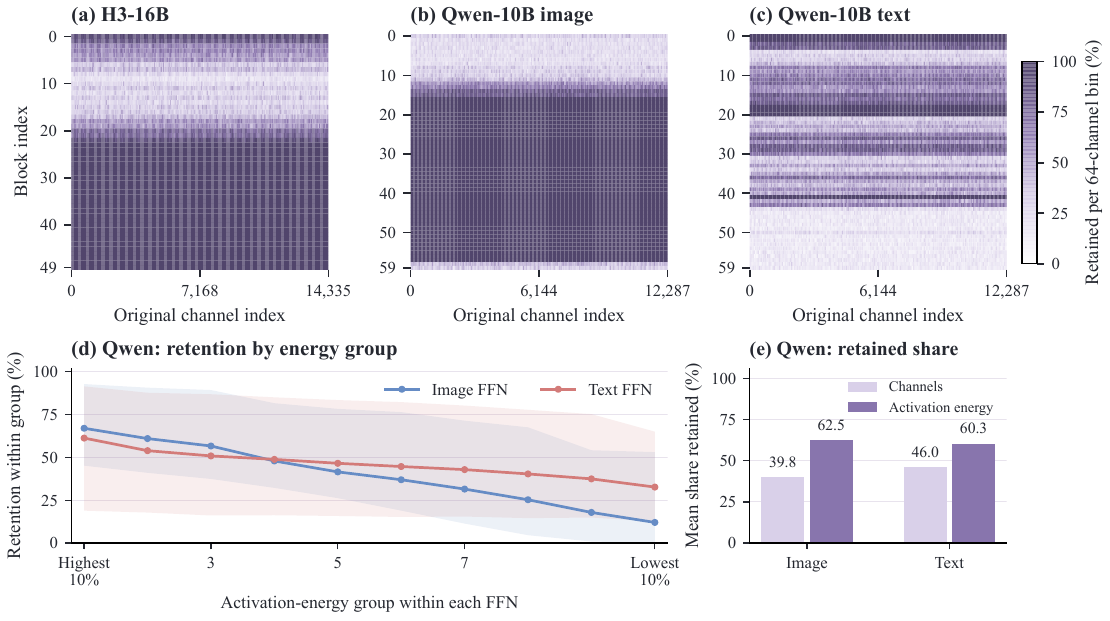}
\caption{FFN channel retention. (a--c) Masks for H3-16B and Qwen-10B image/text FFNs: rows are blocks, columns follow original channel order, and each cell gives the retained percentage among 64 adjacent channels (darker purple means more retained). (d) Retention within activation-energy deciles, highest to lowest, for 18 image and 54 text partially pruned Qwen FFNs. Curves are cross-FFN means; bands span the 10th--90th percentiles. (e) For these same FFNs, mean retained shares of channels (light purple) and activation energy (dark purple).}
\label{supp:fig:ffn-channel-analysis}

\centering
\includegraphics[width=0.82\linewidth,height=0.27\textheight,keepaspectratio]{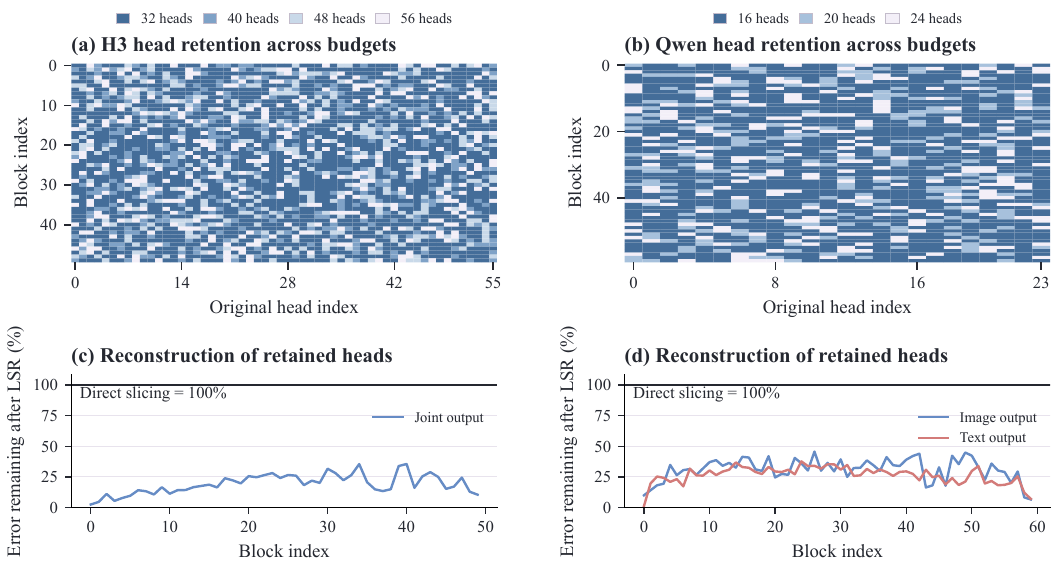}
\caption{Layer-specific head selection and reconstruction. Top: the smallest tested head budget retaining each head ($32,40,48,56$ for H3; $16,20,24$ for Qwen); darker cells survive tighter budgets. Indices are layer-local. Bottom: calibration output-projection MSE after LSR as a percentage of direct-slicing MSE, at 48 heads for H3 and 20 for Qwen. The image and text projections of Qwen are shown separately.}
\label{supp:fig:head-patterns}

\end{figure*}

\paragraph{Within-FFN channel patterns.}
The width $w_j$ determines how many channels remain; the coupled weight-norm score $\rho_{j,i}^{\mathrm{ffn}}$ in Section~\ref{mref:app:candidates} determines which ones. Figure~\ref{supp:fig:ffn-channel-analysis}(a--c) shows these choices in original channel order. Each row is a block, and each cell averages retention over 64 adjacent channels.

To inspect what the ranking preserves, panel (d) groups channels by calibration mean-squared activation after the nonlinearity, from the highest-energy decile to the lowest. Higher-energy groups retain a larger fraction of their channels. For the partially pruned Qwen FFNs (18 image and 54 text), panel (e) shows mean retained channel shares of $39.8\%$ and $46.0\%$, but retained activation-energy shares of $62.5\%$ and $60.3\%$. These per-FFN averages show that the retained channels carry a disproportionate share of activation energy. Energy is used here to analyze the selected channels; the pruning rule remains the weight-norm ranking.

\subsection{Layer-Specific Head Selection and Reconstruction}

At $h_l=48$ on H3 and $h_l=20$ on Qwen, all 50 H3 and 60 Qwen layers use distinct head masks (Figure~\ref{supp:fig:head-patterns}). The nested budget tiers reflect the calibrated ranking in Section~\ref{mref:app:reconstruction}; head indices are layer-local.

LSR lowers calibration MSE relative to direct slicing in all 50 H3 and 120 Qwen image/text output projections. Median error ratios are $18.5\%$ and $29.9\%$, respectively ($18.5$--$20.8\%$ and $29.9$--$30.9\%$ across tested pruned head counts). Refitting therefore corrects much of the slicing error on calibration activations.

\subsection{Initialization Quality Before Recovery Training}
\label{supp:sec:initialization-quality}

\begin{figure*}[!t]
\centering
\includegraphics[width=\textwidth]{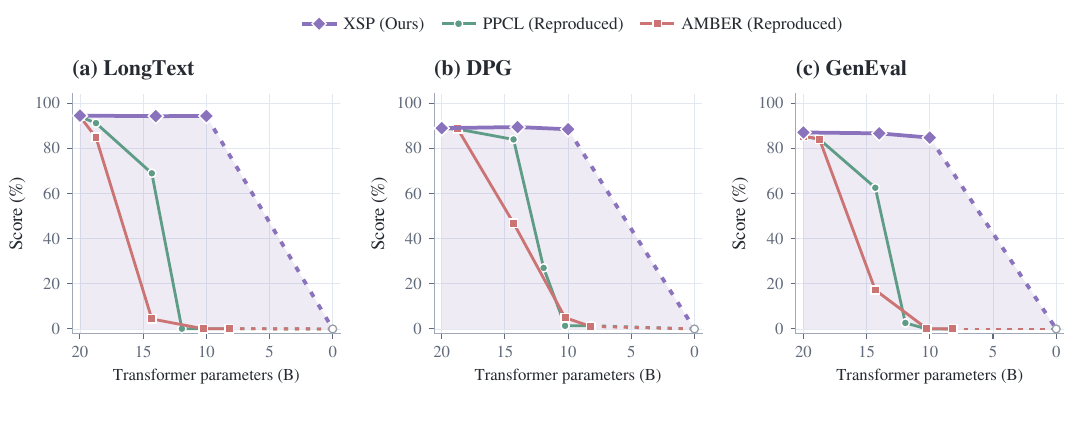}
\caption{\textbf{Initialization quality on Qwen-Image 1.0 before student recovery training.} Panels report (a)~LongText (arithmetic mean of EN and ZH), (b)~DPG, and (c)~GenEval on a 0--100 scale against descending transformer parameters. XSP is plotted at nominal 14B and 10B, and the locally evaluated PPCL depth and AMBER direct-retention initializations at their exact pruned sizes; each series connects to its own source teacher at nominal 20B. LongText uses CFG~4 throughout; DPG and GenEval use CFG~8 for XSP and CFG~4 for depth baselines. Filled markers denote measured checkpoints. Purple shading highlights the XSP curve, connecting lines guide the eye, and dotted tails to the open 0B endpoint are schematic.}
\label{supp:fig:initialization-envelope}
\end{figure*}

The preceding analyses show that XSP retains high-energy FFN channels and reconstructs pruned attention and FFN projections with low local error. Figure~\ref{supp:fig:initialization-envelope} tests how this module-level preservation translates into end-to-end generation quality on Qwen-Image 1.0 before any student recovery training. We compare XSP against locally evaluated PPCL depth and AMBER direct-retention initializations across parameter budgets. XSP applies activation calibration, low-rank AdaLN factorization, and closed-form least-squares reconstruction. Both baselines remove whole transformer blocks under their respective selection rules, with AMBER retaining chosen layers directly without layer averaging.

Both plotted depth-initialization curves drop sharply as the parameter budget shrinks. At $14.313$B, mean LongText accuracy drops to $68.9$ for PPCL depth initialization and $4.4$ for AMBER direct retention; by $10.235$B, both collapse to $0.06$ on LongText, with DPG at $1.37$ and $4.99$ and GenEval at $0.0$ and $0.10$. By contrast, XSP preserves full network depth and compensates removed capacity within each block. It sustains a LongText mean of $94.2$ at 14B and $94.3$ at 10B (versus $94.45$ for its 20B teacher), alongside $88.4$ DPG and $84.7$ GenEval at 10B. Retaining high generation quality at initialization makes student recovery training an optional refinement rather than a prerequisite for usable synthesis, motivating the recovery experiments in Section~\ref{supp:extreme_prune_ratio_tiny_prune}.

\section{Compatibility to Recovery Training For Extreme Pruning}
\label{supp:extreme_prune_ratio_tiny_prune}
The main text has shown results of 30-50\% pruning ratio. Without costly gradient-based training, XSP outperforms existing methods. In Table~\ref{tab:extreme_pruning}, we exhibit results under extreme pruning towards a native tiny model Z-Image Turbo.
\begin{table*}[t]
\caption{
Results towards compact tiny model and the compatibility to recovery training. We retrieve the official released model of Z-Image Turbo and evaluate locally. For TMP, we directly report their paper's results.  TMP-lite$^{\dagger}$: We reproduce and apply only the first stage of recovery training used by TMP and apply it to our XSP-pruned model.
}
\label{tab:extreme_pruning}
\centering
\renewcommand{\arraystretch}{1.12}
\setlength{\tabcolsep}{4pt}
\resizebox{\textwidth}{!}{
\begin{tabular}{c|c|c|c|c|cccc}
\hline
\multirow{2}{*}{Models}
& \multirow{2}{*}{Methods}
& \multirow{2}{*}{Params. (B)}
& \multirow{2}{*}{KD Recovery}
& \multirow{2}{*}{Post-KD FT}
& \multicolumn{2}{c}{LongText$\uparrow$}
& \multicolumn{2}{c}{OneIG$\uparrow$} \\
\cline{6-9}
 & & & & & EN & ZH & EN & ZH \\
\hline
\multirow{4}{*}{Z-Image Turbo}
& Base model & 6.2 & - & - & 0.921 & 0.933 & 0.523 & 0.498 \\
\cline{2-9}
& TMP & 4.6 & \ding{52} & \ding{56}
& \underline{0.889} & \underline{0.882} & \textbf{0.530} & \textbf{0.500} \\
& \textbf{XSP} (Ours) & 4.6 & \ding{56} & \ding{56}
& 0.739 & 0.793 & 0.477 & 0.441 \\
& \textbf{XSP-TMP-lite}$^{\dagger}$ (Ours) & 4.6 & \ding{52} & \ding{56}
& \textbf{0.908} & \textbf{0.902} & \underline{0.503} & \underline{0.479} \\
\hline
\end{tabular}
}
\end{table*}

Actually, Z-Image Turbo is a native compact model that does not necessitate pruning. We provide its case study to verify the generality of our method. In a previous work TMP, Zhang \textit{et al}. proposes to prune FFN of Z-Image Turbo by 33\%, leading to an overall 25\% pruned DiT model (4.65B). We follow exactly the same parameter amount reduction but we differ them in heterogeneous FFNs pruning by our proposed MCKP. 

As an efficient framework that eliminates the need for heavy gradient-descent-based retraining, XSP should not be held to the same performance expectations as fine-tuned counterparts, particularly under extreme pruning regimes. Nevertheless, this also provides a timely opportunity to demonstrate XSP's compatibility with recovery training algorithms, showing that it can serve as a strong initialization even under extreme pruning. We partially reproduce the recovery training proposed by TMP. For simplicity, we implement only the off-policy feature distillation since there lacks disclosed details about how to hybrid the on-policy version in their technical report. As a result, XSP-TMP-lite variant still achieves promising results with such TMP-lite KD for only 2k iterations of distillation with 128$\times$H20. This demonstrate the potential of XSP serving as a strong pruning initialization under extreme pruning. 
\section{Compatibility with Quantization}
\label{supp:sec:quantization}

Structural pruning reduces network dimensions, while post-training quantization (PTQ) lowers the precision of the remaining weights and activations. We combine the two by quantizing the primary attention and FFN linear layers of XSP-H3-16B, without additional fine-tuning or distillation.

We use ConvRot in ComfyUI's \texttt{comfy-kitchen} for W8A8 (8-bit weights and activations) and W4A8 (4-bit weights, 8-bit activations); other modules retain their existing precision. Memory is measured during the DiT forwards of a $3360\times5024$ six-step workload on an NVIDIA L20Y. The full DiT stays GPU-resident, while the text encoder and VAE are CPU-offloaded.

\begin{center}
\begin{minipage}{\columnwidth}
\centering
\captionof{table}{Peak PyTorch-allocated GPU memory during DiT forwards, under the workload and placement described in Section~\ref{supp:sec:quantization}.}
\label{supp:tab:quantization-memory}
\small
\setlength{\tabcolsep}{5pt}
\renewcommand{\arraystretch}{1.08}
\begin{tabular*}{\linewidth}{@{\extracolsep{\fill}}lrr@{}}
\toprule
Precision & Peak memory (GiB) & Reduction \\
\midrule
BF16 & 33.61 & -- \\
W8A8 & 19.81 & 41.1\% \\
W4A8 & 13.77 & 59.0\% \\
\bottomrule
\end{tabular*}
\end{minipage}
\end{center}

Table~\ref{supp:tab:quantization-memory} shows that peak PyTorch-allocated memory falls from $33.61$ GiB in BF16 to $19.81$ GiB with W8A8 and $13.77$ GiB with W4A8, reductions of $41.1\%$ and $59.0\%$ on this workload. Figure~\ref{supp:fig:quantization} compares the same pruned model across these precisions. The harbour scene and poster retain similar overall layouts and recognizable subjects, while local details vary.

\section{Qualitative Comparisons Across Compression Scales}
\label{supp:sec:qualitative}
Examples follow Sections~\ref{supp:sec:t2i-details}, \ref{supp:sec:editing}, and~\ref{supp:sec:quantization}: T2I, editing, then quantization. Comparisons within each family use matched prompts, seeds, resolution, and sampling settings, with original aspect ratios.

\paragraph{Text-to-image generation.}
Figures~\ref{supp:fig:h3-denseocr}--\ref{supp:fig:h3-nature} compare original H3-33B with XSP-20B/16B; Figures~\ref{supp:fig:qwen-lifestyle} and~\ref{supp:fig:qwen-nature} compare original Qwen-20B with XSP-14B/10B. H3 uses shared official PE at approximately 16 MP and six steps; Qwen uses original prompts at approximately 1.7 MP and 50 steps. Posters, portraits, and natural scenes expose typography, composition, and fine detail.

\paragraph{Image editing.}
Figures~\ref{supp:fig:h3-editing} and~\ref{supp:fig:h3-editing-multiref} pair the reference inputs with original H3-33B and pruned XSP-16B outputs, showing both single-reference changes and combinations of two to five inputs.

\paragraph{Quantization.}
Figure~\ref{supp:fig:quantization} compares BF16, W8A8, and W4A8 versions of XSP-H3-16B at a fixed model size.

\phantomsection\label{supp:sec:h3}
\begin{figure*}[p]
\centering
\includegraphics[width=\linewidth,height=0.77\textheight,keepaspectratio]{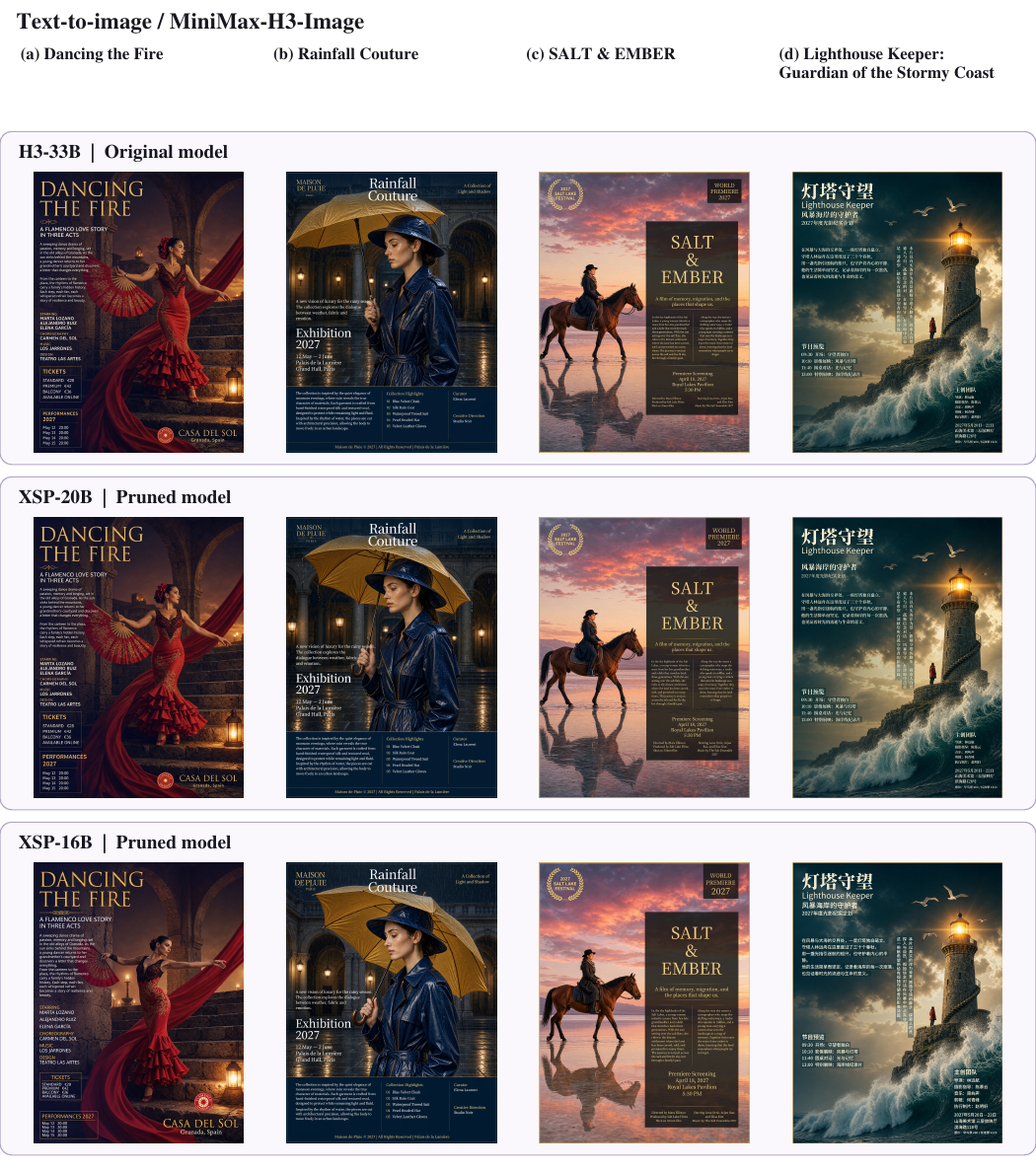}
\caption{Dense-text posters on MiniMax-H3-Image. Each column is a prompt; rows show original H3-33B, XSP-20B (rank-16 AdaLN), and XSP-16B (48-head attention and MCKP FFNs). The posters expose text layout and typography alongside the subjects and backgrounds.}
\label{supp:fig:h3-denseocr}
\end{figure*}
\begin{figure*}[p]
\centering
\includegraphics[width=\linewidth]{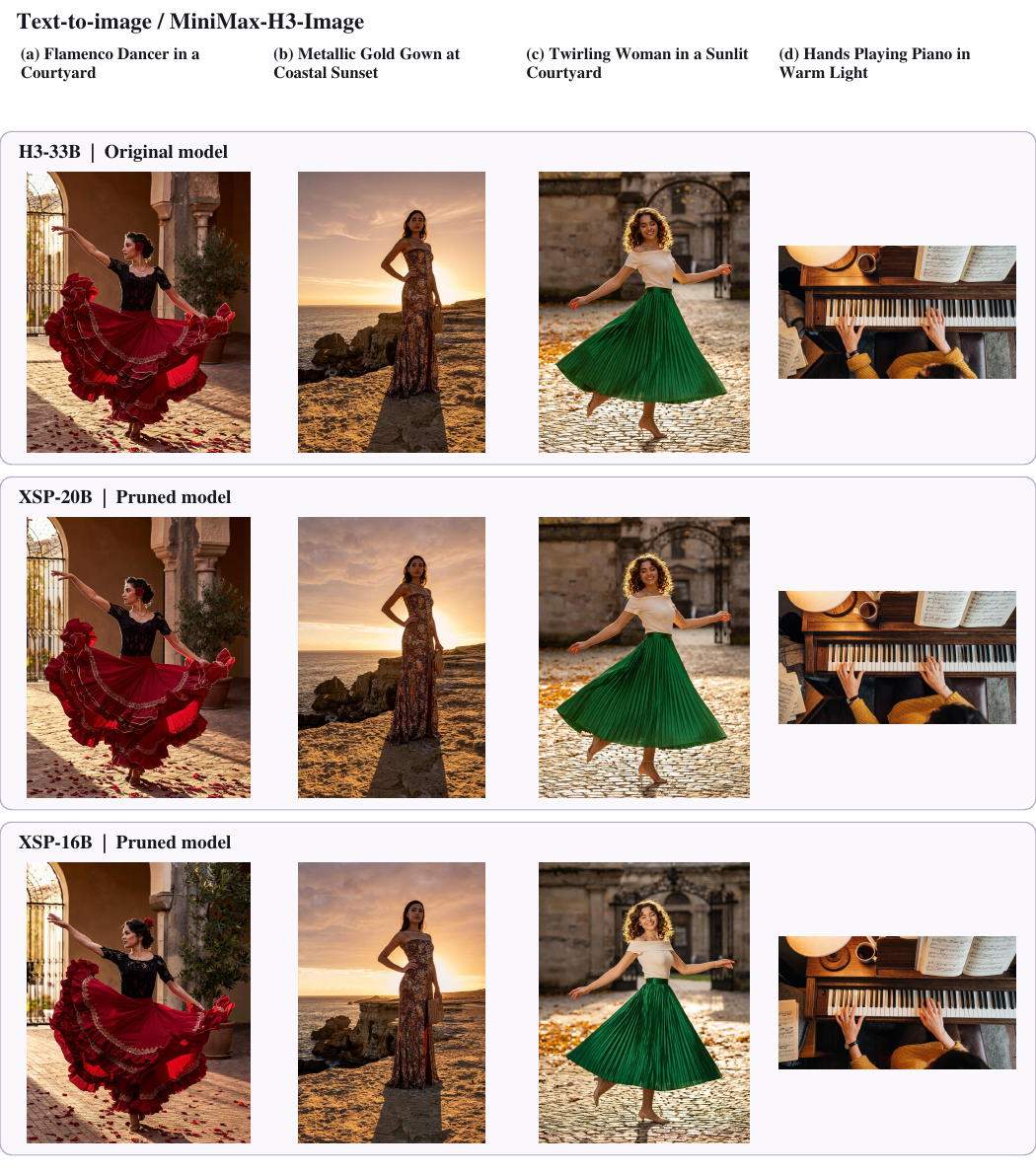}
\caption{Portrait and human-subject examples on MiniMax-H3-Image. Each column is a prompt; rows show original H3-33B, XSP-20B, and XSP-16B. The comparison shows pose, clothing, and scene composition across model sizes.}
\label{supp:fig:h3-people}
\end{figure*}
\begin{figure*}[p]
\centering
\includegraphics[width=\linewidth]{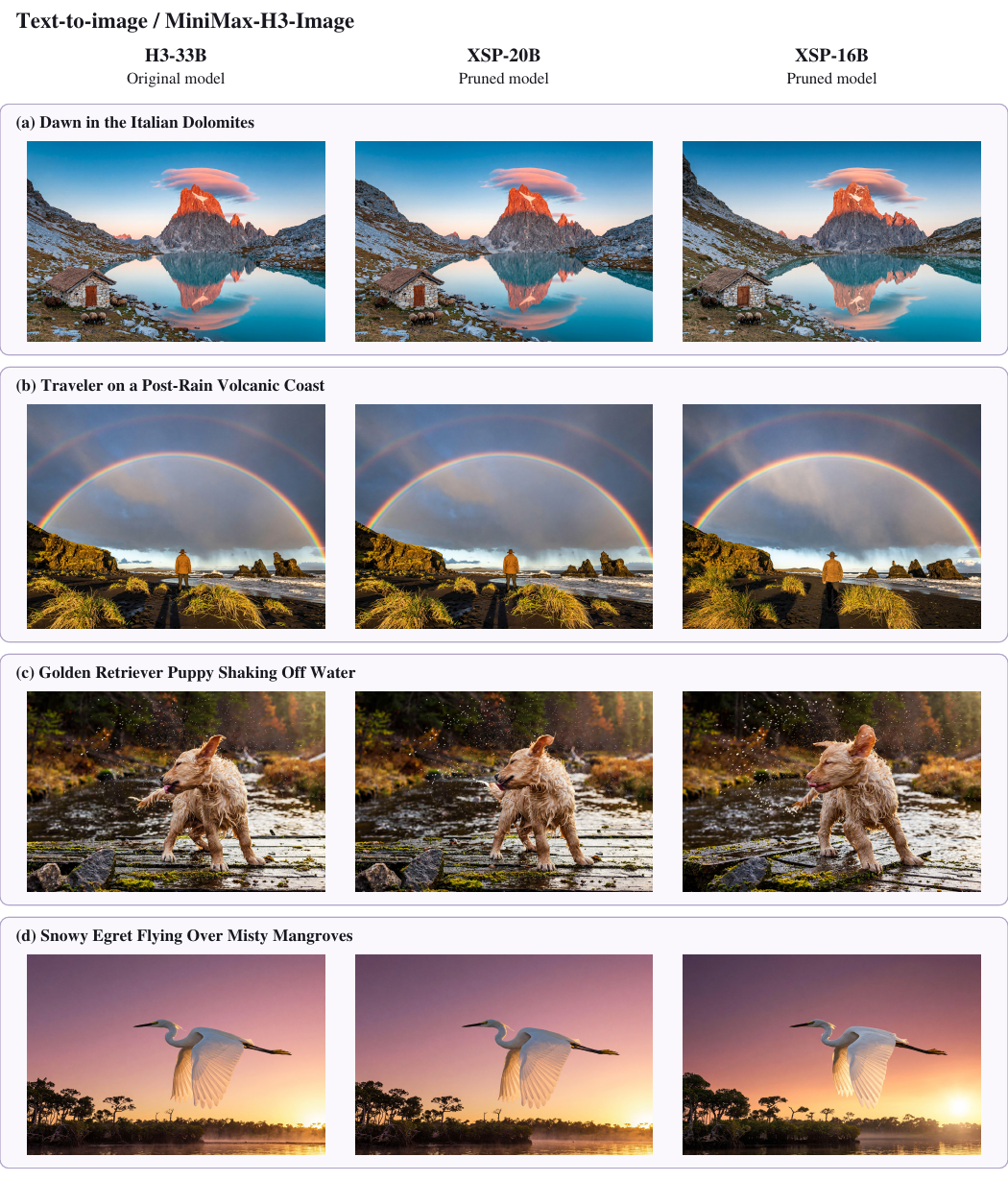}
\caption{Landscape and wildlife examples on MiniMax-H3-Image. Columns show original H3-33B, XSP-20B, and XSP-16B from left to right, highlighting scene layout and natural detail across model sizes.}
\label{supp:fig:h3-nature}
\end{figure*}
\phantomsection\label{supp:sec:qwen}
\begin{figure*}[p]
\centering
\includegraphics[width=\linewidth]{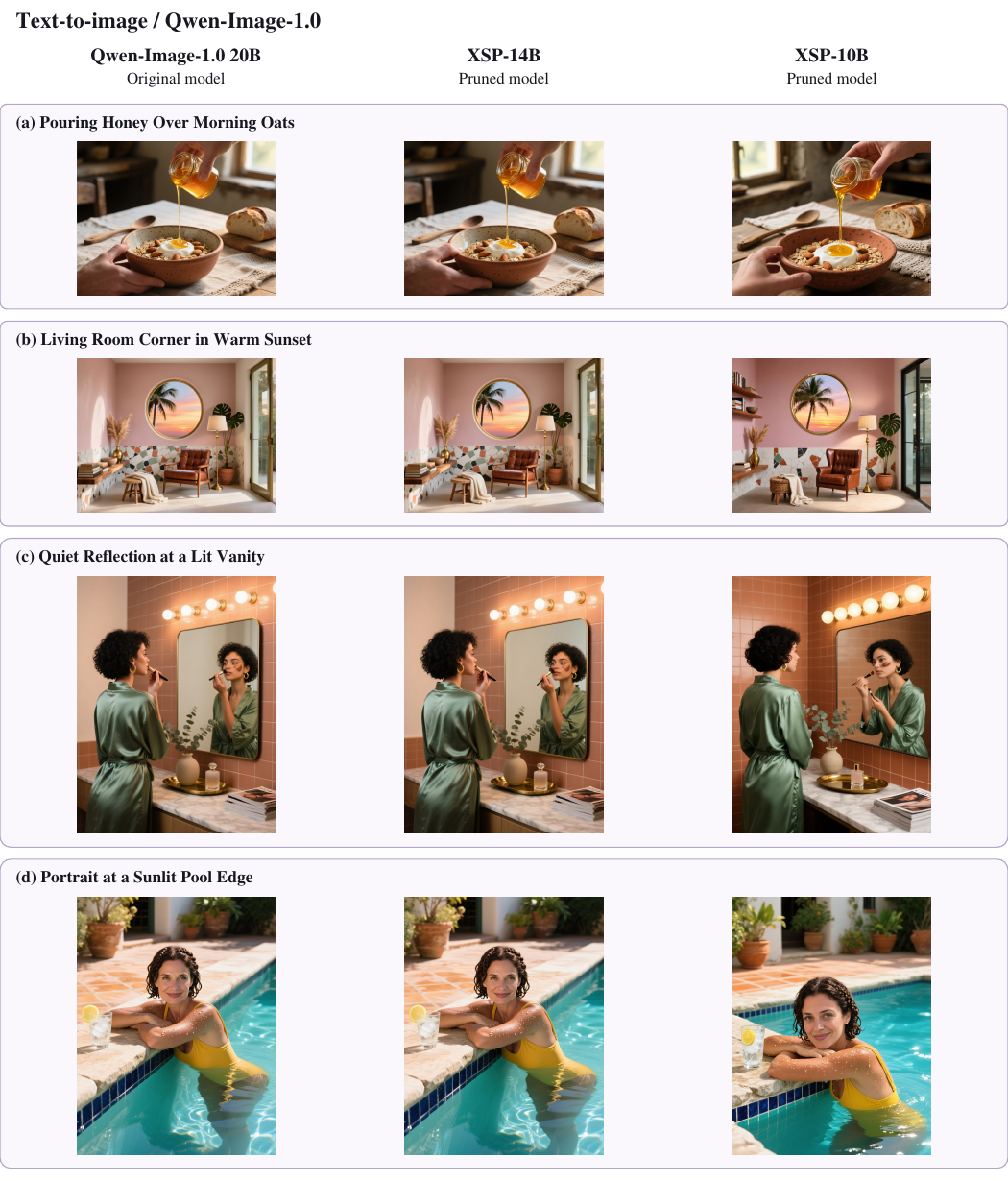}
\caption{Photographic lifestyle and portrait examples on Qwen-Image-1.0. Columns show original 20B, XSP-14B (rank-48 AdaLN), and XSP-10B (20-head attention and MCKP FFNs), from left to right.}
\label{supp:fig:qwen-lifestyle}
\end{figure*}
\begin{figure*}[p]
\centering
\includegraphics[width=\linewidth]{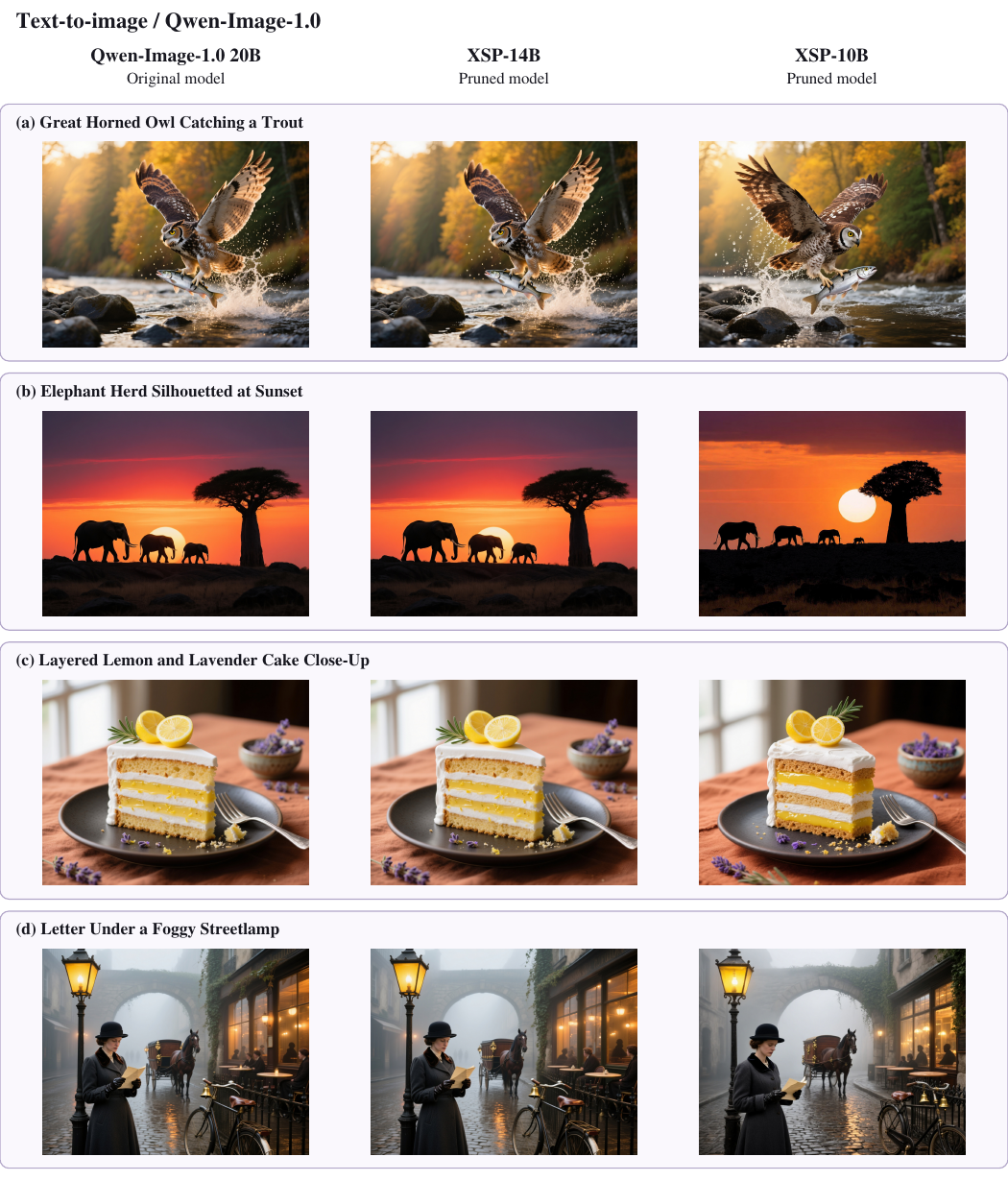}
\caption{Natural scenes on Qwen-Image-1.0. Columns show original 20B, XSP-14B, and XSP-10B from left to right, allowing comparison of subject appearance, lighting, and background detail.}
\label{supp:fig:qwen-nature}
\end{figure*}
\begin{figure*}[p]
\centering
\includegraphics[width=0.97\linewidth]{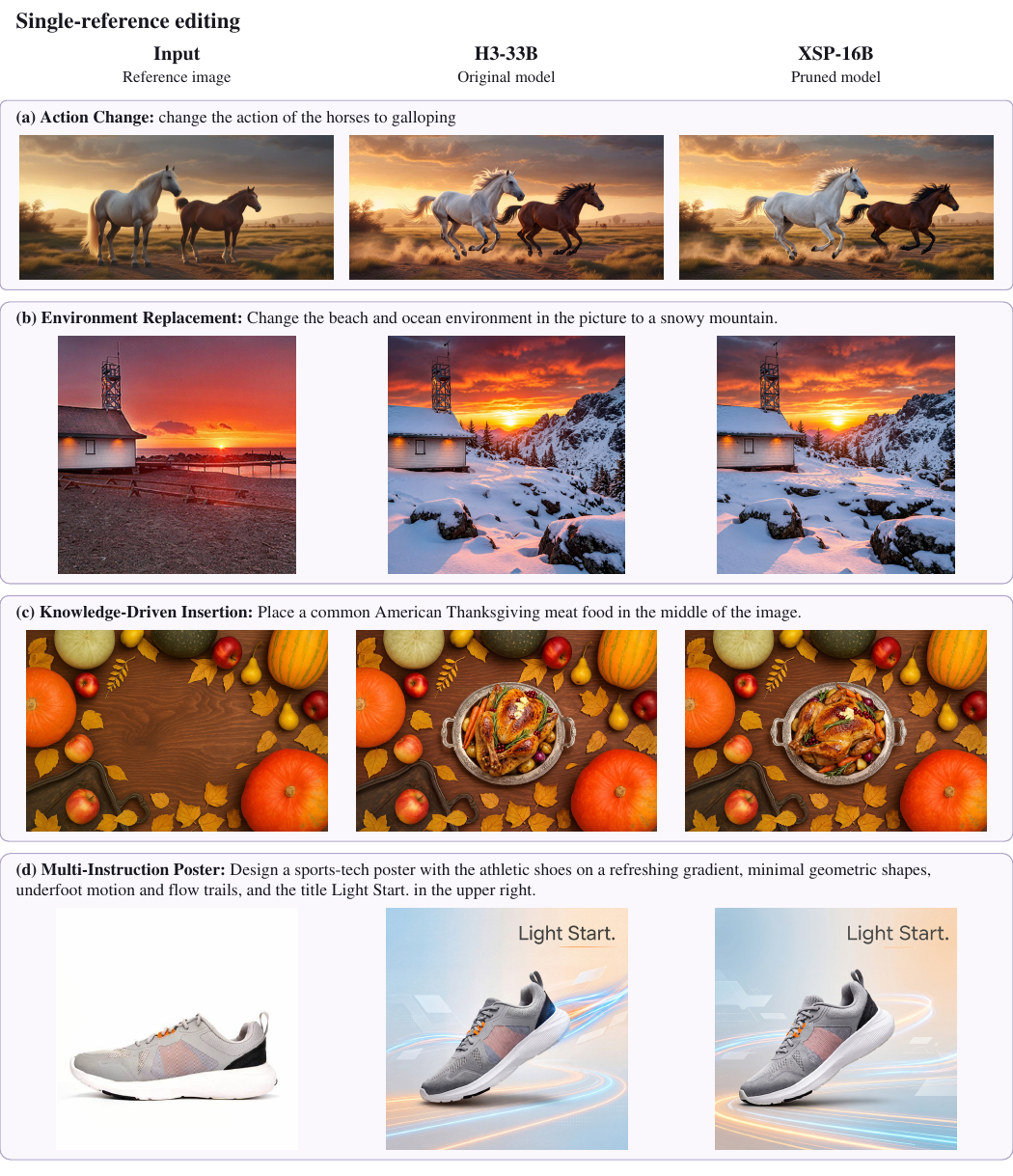}
\caption{Single-reference image editing. Columns show the input image, original H3-33B output, and pruned XSP-16B output. The requested change is shown for each row; row (d)'s instruction is shortened for display.}
\label{supp:fig:h3-editing}
\end{figure*}
\begin{figure*}[p]
\centering
\includegraphics[width=\linewidth]{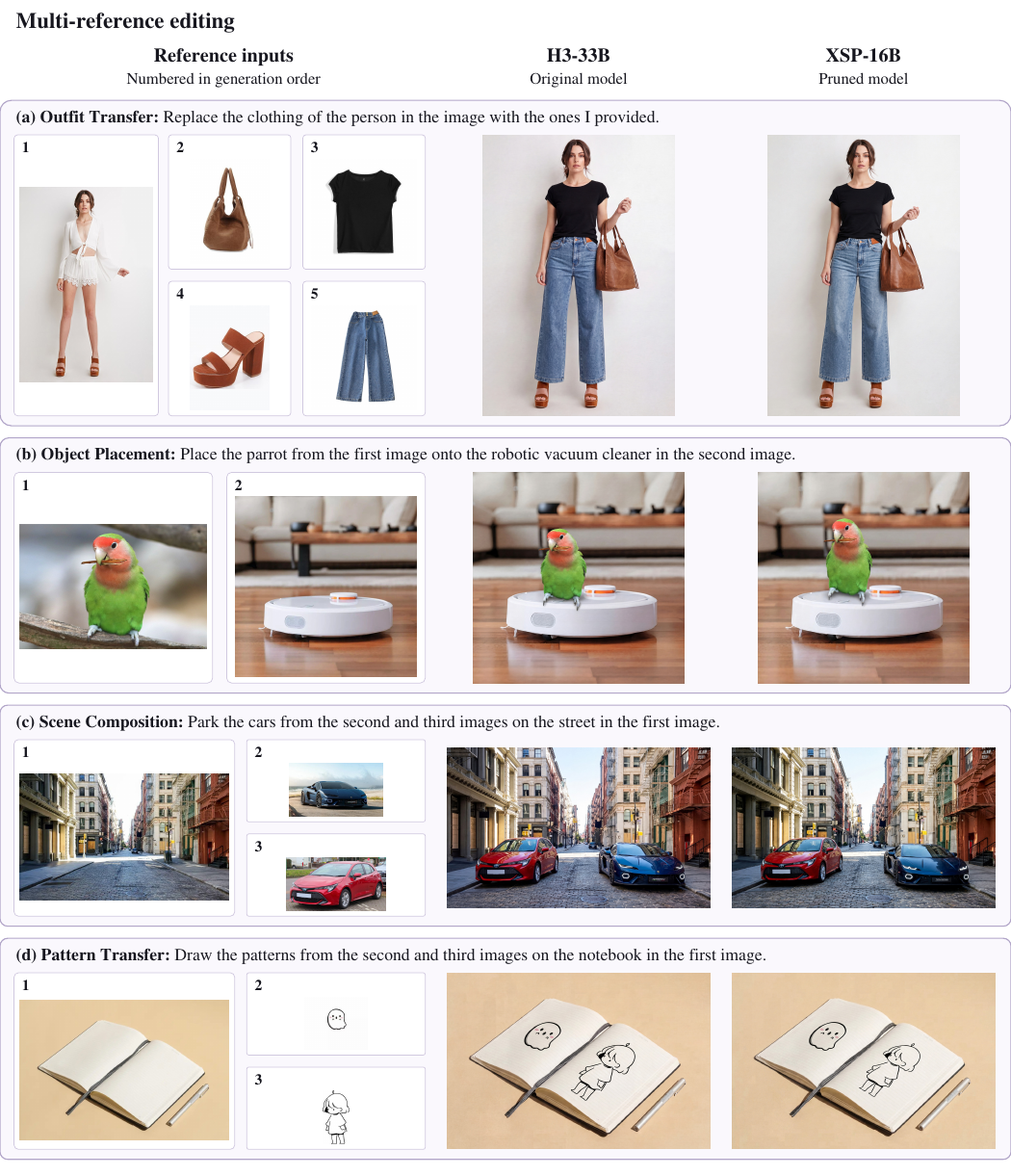}
\caption{Multi-reference image editing with two to five inputs, numbered in generation order. Original H3-33B and pruned XSP-16B receive the same ordered references and instruction; the output comparison shows how each model combines the supplied content.}
\label{supp:fig:h3-editing-multiref}
\end{figure*}
\begin{figure*}[p]
\centering
\includegraphics[width=0.76\linewidth]{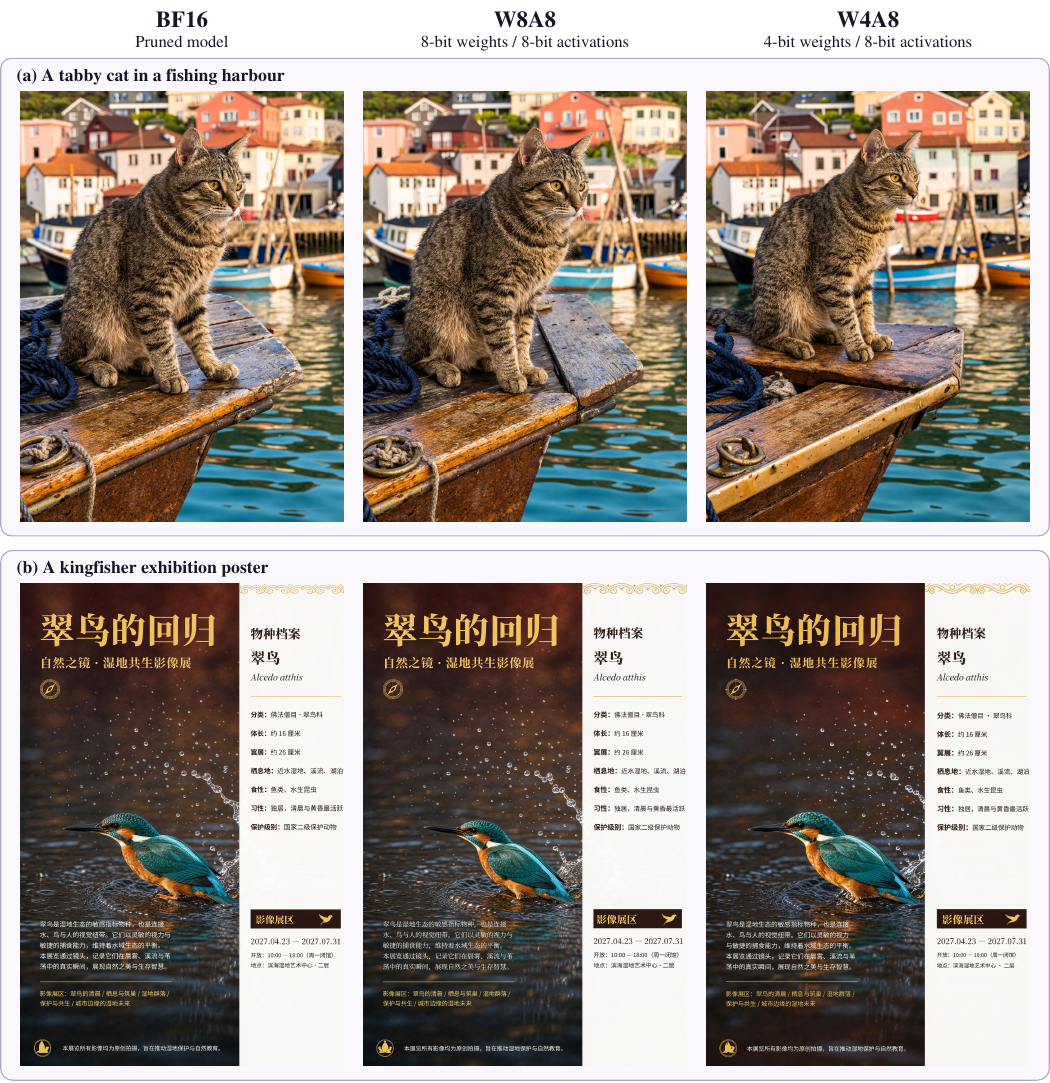}
\caption{Post-training quantization of XSP-H3-16B: BF16, W8A8, and W4A8 from left to right. The harbour cat and Chinese exhibition poster are generated at native 4K resolution with matched PE-expanded prompts, initial noise, and sampling settings. The comparison complements the memory measurements in Section~\ref{supp:sec:quantization}.}
\label{supp:fig:quantization}
\end{figure*}

\clearpage
\endgroup

\end{document}